\documentclass[runningheads]{llncs}
\usepackage{graphicx}

\usepackage{tikz}
\usepackage{comment}
\usepackage{amsmath,amssymb} %
\usepackage{color}
\usepackage{multirow}
\usepackage{hyperref}
\usepackage{bbm}
\usepackage{xspace}
\usepackage{wrapfig}
\usepackage[caption=false]{subfig}
\graphicspath{{figure/}}

\newcommand{\ie}{\textit{i}.\textit{e}.,\xspace}
\newcommand{\eg}{\textit{e}.\textit{g}.,\xspace}
\newcommand{\et}{\textit{et al}.\xspace}
\DeclareMathOperator*{\argmax}{arg\,max}

\newcommand{\squishlist}{
	\begin{list}{$\bullet$}
		{ \setlength{\itemsep}{0pt}
			\setlength{\parsep}{1pt}
			\setlength{\topsep}{1pt}
			\setlength{\partopsep}{0pt}
			\setlength{\leftmargin}{1.5em}
			\setlength{\labelwidth}{1em}
			\setlength{\labelsep}{0.5em} } }
\newcommand{\squishend}{\end{list}
}

\usepackage[accsupp]{axessibility}  %

\begin{document}
\pagestyle{headings}
\mainmatter
\def\ECCVSubNumber{7278}  %

\title{ActionFormer: Localizing Moments of Actions with Transformers} %

\titlerunning{ActionFormer: Localizing Moments of Actions with Transformers}
\author{{Chen-Lin Zhang\thanks{Work was done when visiting UW Madison.}}\inst{1,2} \and
Jianxin Wu\inst{1} \and
{Yin Li}\inst{3}}
\authorrunning{Zhang et al.}
\institute{State Key Laboratory for Novel Software Technology, Nanjing University, China \and
4Paradigm Inc, Beijing, China \and
University of Wisconsin-Madison, USA \\
\email{\{zclnjucs, wujx2001\}@gmail.com \qquad yin.li@wisc.edu}}
\maketitle

\begin{abstract}
Self-attention based Transformer models have demonstrated impressive results for image classification and object detection, and more recently for video understanding. Inspired by this success, we investigate the application of Transformer networks for temporal action localization in videos. To this end, we present ActionFormer---a simple yet powerful model to identify actions in time and recognize their categories in a single shot, without using action proposals or relying on pre-defined anchor windows. ActionFormer combines a multiscale feature representation with local self-attention, and uses a light-weighted decoder to classify every moment in time and estimate the corresponding action boundaries. We show that this orchestrated design results in major improvements upon prior works. Without bells and whistles, ActionFormer achieves 71.0\% mAP at tIoU$=$0.5 on THUMOS14, outperforming the best prior model by 14.1 absolute percentage points. Further, ActionFormer demonstrates strong results on ActivityNet 1.3 (36.6\% average mAP) and EPIC-Kitchens 100 (+13.5\% average mAP over prior works). Our code is available at \url{https://github.com/happyharrycn/actionformer_release}. 
\end{abstract}

\section{Introduction}
\label{sec:intro}

Identifying action instances in time and recognizing their categories, known as temporal action localization (TAL), remains a challenging problem in video understanding. Significant progress has been made in developing deep models for TAL. Most previous works have considered using action proposals~\cite{bmniccv2019} or anchor windows~\cite{gaussiancvpr2019}, and developed convolutional~\cite{ssniccv2017,cdc2017cvpr}, recurrent~\cite{sstadbmvc2017}, and graph~\cite{bcgnneccv2020,gtadcvpr2020,yang2022acgnet} neural networks for TAL. Despite a steady progress on major benchmarks, the accuracy of existing methods usually comes at a price of modeling complexity, with increasingly sophisticated proposal generation, anchor design, loss function, network architecture, and output decoding process.

\noindent\begin{wrapfigure}{l}{0.52\textwidth} 
	\centering
	\includegraphics[width=0.95\linewidth]{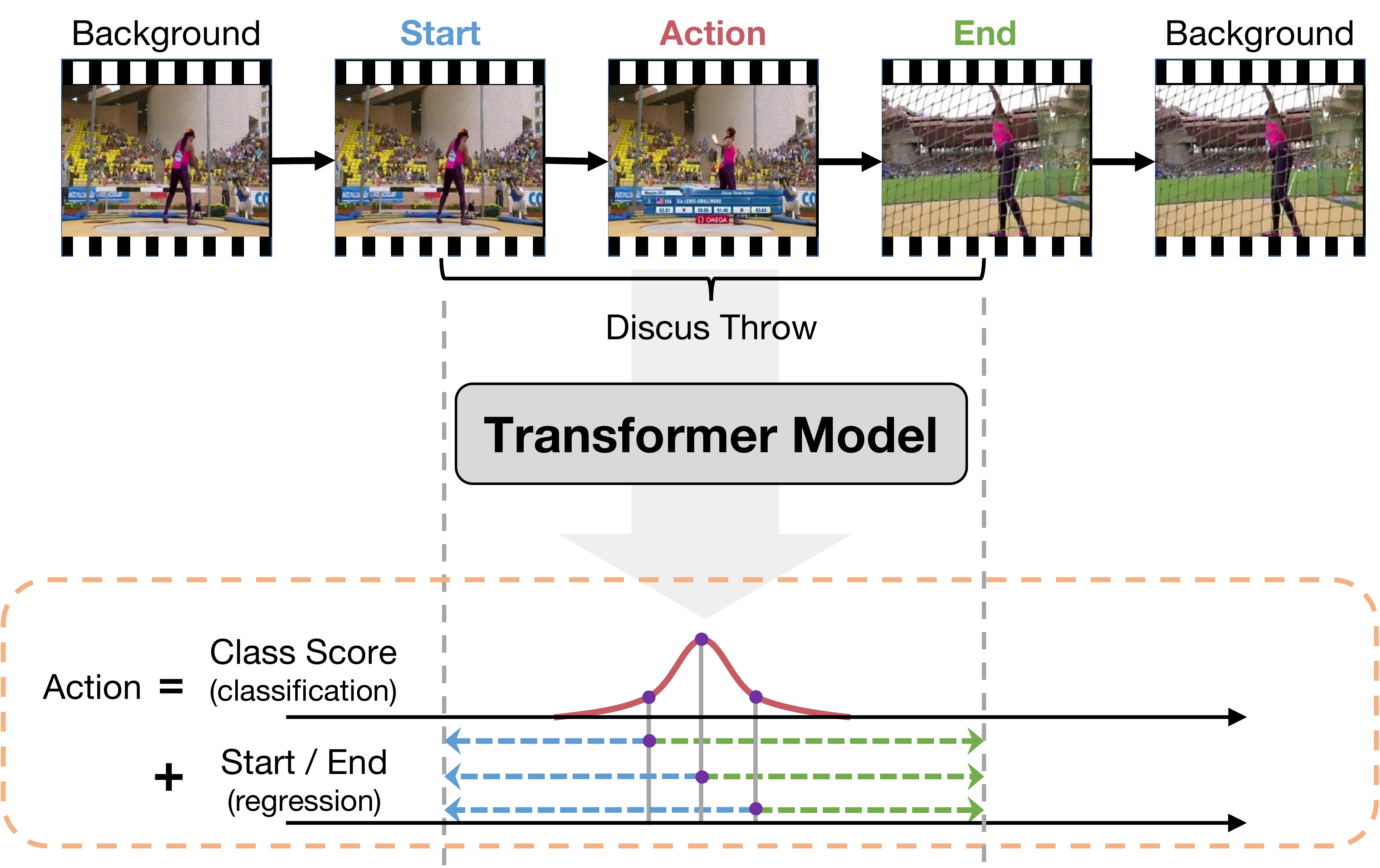}
	\caption{\textbf{An illustration of our ActionFormer}. We propose a Transformer based model to localize action instances in time (\emph{top}) by (1) classifying every moment into action categories and (2) estimating their distances to action boundaries (\emph{bottom}).}
	\label{fig:teaser}
\end{wrapfigure}%
In this paper, we adopt a minimalist design and develop a Transformer based model for TAL, inspired by the recent success of Transformers in NLP~\cite{transformernips2017,devlin2019bert} and vision~\cite{vitarxiv2020,swiniccv2021,detreccv2020}. Originally developed for sequence data, Transformers use self-attention to model long-range dependencies, and thus are a natural fit for TAL in untrimmed videos. Our method, illustrated in Fig.\ \ref{fig:teaser}, adapts local self-attention to model temporal context in an input untrimmed videos, classifies every moment, and regresses their corresponding action boundaries. The result is a deep model trained using standard classification and regression loss, and can localize moments of actions in a single shot, without using action proposals or pre-defined anchor windows. 

Specifically, our model, dubbed ActionFormer, integrates local self-attention to extract a feature pyramid from an input video. Each location in the output pyramid represents a moment in the video, and is treated as an action candidate. A lightweight convolutional decoder is further employed on the feature pyramid to classify these candidates into foreground action categories, and to regress the distance between a foreground candidate and its action onset and offset. The results can be easily decoded into actions with their labels and temporal boundaries. Our method thus provides a {\it single-stage anchor-free} model for TAL. 

We show that such a simple model, with proper design, can be surprisingly powerful for TAL. In particular, ActionFormer establishes a new state of the art across several major TAL benchmarks, surpassing previous works by a significant margin. For example, ActionFormer achieves 71.0\% \textit{m}AP at tIoU$=$0.5 on THUMOS14, outperforming the best prior model by 14.1 absolute percentage points. Further, ActionFormer reaches an average \textit{m}AP of 36.6\% on ActivityNet 1.3. More importantly, ActionFormer shows impressive results on EPIC-Kitchens 100 for egocentric action localization, with a boost of over 13.5 absolute percentage points in average \textit{m}AP. %

Our work is based on simple techniques, supported by favourable empirical results, and validated by extensive ablation experiments, at our best. Our main contributions are summarized as follows. First, we are among the first to propose a Transformer based model for single-stage anchor-free TAL. Second, we study key design choices of developing Transformer models for TAL, and demonstrate a simple model that works surprisingly well. Finally, our model achieves state-of-the-art results across major benchmarks and offers a solid baseline for TAL.

\section{Related Works}
\label{sec:related_work}
\noindent \textbf{Two-stage TAL}. These approaches first generate candidate video segments as action proposals, and further classify the proposals into actions and refine their temporal boundaries. Previous works focused on action proposal generation, by either classifying anchor windows~\cite{caba2016fast,escorcia2016daps,sstcvpr2017} or detecting action boundaries~\cite{bsneccv2018,bmniccv2019,mggcvpr2019,gong2020scale,bottumuptaleccv2020}, and more recently using a graph representation~\cite{bcgnneccv2020,gtadcvpr2020} or Transformers~\cite{tan2021relaxed,chang2021augmented,wang2021temporal}. Others have integrated proposal generation and classification into a single model~\cite{shou2016temporal,ssniccv2017,cdc2017cvpr,fasterrcnncvpr2018}. More recent effort investigates the modeling of temporal context among proposals using graph neural networks~\cite{pgcniccv2019,gtadcvpr2020,vsgniccv2021} or attention and self-attention mechanisms~\cite{contextlociccv2021,tcanetcvpr2021,csaiccv2021}. Similar to previous approaches, our method considers the modeling of long-term temporal context, yet uses a self-attention within a Transformer model. Different from previous approaches, our model detects actions without using proposals.\smallskip

\noindent \textbf{Single-stage TAL}. Several recent works focused on single-stage TAL, seeking to localize actions in a single shot without using action proposals. Many of them are anchor-based (\eg using anchor windows sampled from sliding windows). Lin \et\ \cite{sstadmm2017} presented the first single-stage TAL using convolutional networks, borrowing ideas from a single-stage object detector~\cite{liu2016ssd}. Buch \et\ \cite{sstadbmvc2017} presented a recurrent memory module for single-stage TAL. Long \et\  \cite{gaussiancvpr2019} proposed to use Gaussian kernels to dynamically optimize the scale of each anchor, based on a 1D convolutional network. Yang \et\ \cite{a2nettip2020} explored the combination of anchor-based and anchor-free models for single-stage TAL, again using convolutional networks. More recently, Lin \et\ \cite{afsdcvpr2021} proposed an anchor-free single-stage model by designing a saliency-based refinement module incorporated in convolutional network. Similar ideas were also explored in video grounding~\cite{zeng2020dense}. 

Our model falls into the category of single-stage TAL. Indeed, our formulation follows a minimalist design of sequence labeling by classifying every moment and regressing their action boundaries, previously discussed in~\cite{a2nettip2020,afsdcvpr2021}. The key difference is that we design a Transformer network for action localization. The result is a single stage anchor-free model that outperforms all previous methods. A concurrent work from Liu \et\ \cite{tadtrarxiv2021} also used Transformer for TAL, yet considered a set prediction problem similar to DETR~\cite{detreccv2020}. %
\smallskip

\noindent \textbf{Spatial-temporal Action Localization}.
A related yet different task, known as spatial-temporal action localization, is to detect the actions both temporally and spatially, in the form of moving bounding boxes of an actor. It is possible that TAL might be used as a first step for spatial-temporal localization. Girdhar \et\ \cite{actiontransformercvpr2019} proposed to use Transformer for spatial-temporal action localization. While both our work and~\cite{actiontransformercvpr2019} use Transformer, the two models differ significantly. We consider a sequence of video frames as the inputs, while~\cite{actiontransformercvpr2019} used a set of 2D object proposals. Moreover, our work addresses a different task of TAL.\smallskip

\noindent \textbf{Object Detection}. TAL models have been heavily influenced by the developments of object detection models. Some of our model design, including the multiscale feature representation and convolutional decoder, is inspired by feature pyramid network~\cite{fpncvpr2017} and RetinaNet~\cite{focaliccv2017}. Our training using center sampling also stems from recent single-stage object detectors~\cite{duan2019centernet,fcosiccv2019,zhang2020bridging}.\smallskip

\noindent \textbf{Vision Transformer}. Transformer models were originally developed for NLP tasks~\cite{transformernips2017}, and has demonstrated recent success for many vision tasks. ViT~\cite{vitarxiv2020} presented the first pure Transformer-based model that can achieve state-of-the-art performances on image classification. Subsequent works, including DeiT~\cite{deiticml2021}, T2T-ViT~\cite{t2tviticcv2021}, Swin Transformer~\cite{swiniccv2021}, Focal Transformer~\cite{yang2021focal} and PVT~\cite{pvticcv2021}, have further pushed the envelope, resulting in vision Transformer backbones with impressive results on classification, segmentation, and detection tasks. Transformer have also been explored in object detection~\cite{detreccv2020,deformabledetriclr2021,dynamicdetriccv2021,pnpiccv2021}, semantic segmentation~\cite{transsegcvpr2021,xie2021segformer,cheng2021per}, and video representation learning~\cite{arnab2021vivit,liu2021video,fan2021multiscale}. Our model builds on these developments and presents one of the first Transformer models for TAL.  

\section{ActionFormer: A Simple Transformer Model for Temporal Action Localization}
\label{sec:method}
Given an input video $\mathbf{X}$, we assume that $\mathbf{X}$ can be represented using a set of feature vectors $\mathbf{X}=\{\mathbf{x}_1, \mathbf{x}_2, \ldots, \mathbf{x}_T\}$ defined on discretized time steps $t=\{1, 2, \ldots, T\}$, where the total duration $T$ varies across videos. For example, $\mathbf{x}_t$ can be the feature vector of a video clip at moment $t$ extracted from a 3D convolutional network. The goal of temporal action localization is to predict the action label $\mathbf{Y}=\{\mathbf{y}_1, \mathbf{y}_2, \ldots, \mathbf{y}_N\}$ based on the input video sequence $\mathbf{X}$. $\mathbf{Y}$ consists of $N$ action instances $\mathbf{y}_i$, where $N$ also varies across videos. Each instance $\mathbf{y}_i=(s_i, e_i, a_i)$ is defined by its starting time $s_i$ (onset), ending time $e_i$ (offset) and its action label $a_i$, where $s_i \in [1, T]$, $e_i \in [1, T]$, $a_i \in \{1,.., C\}$ ($C$ pre-defined categories) and $s_i < e_i$. The task of TAL is thus a challenging problem of structured output prediction.

\begin{figure*}[t!]
	\centering 
	\includegraphics[width=0.9\linewidth]{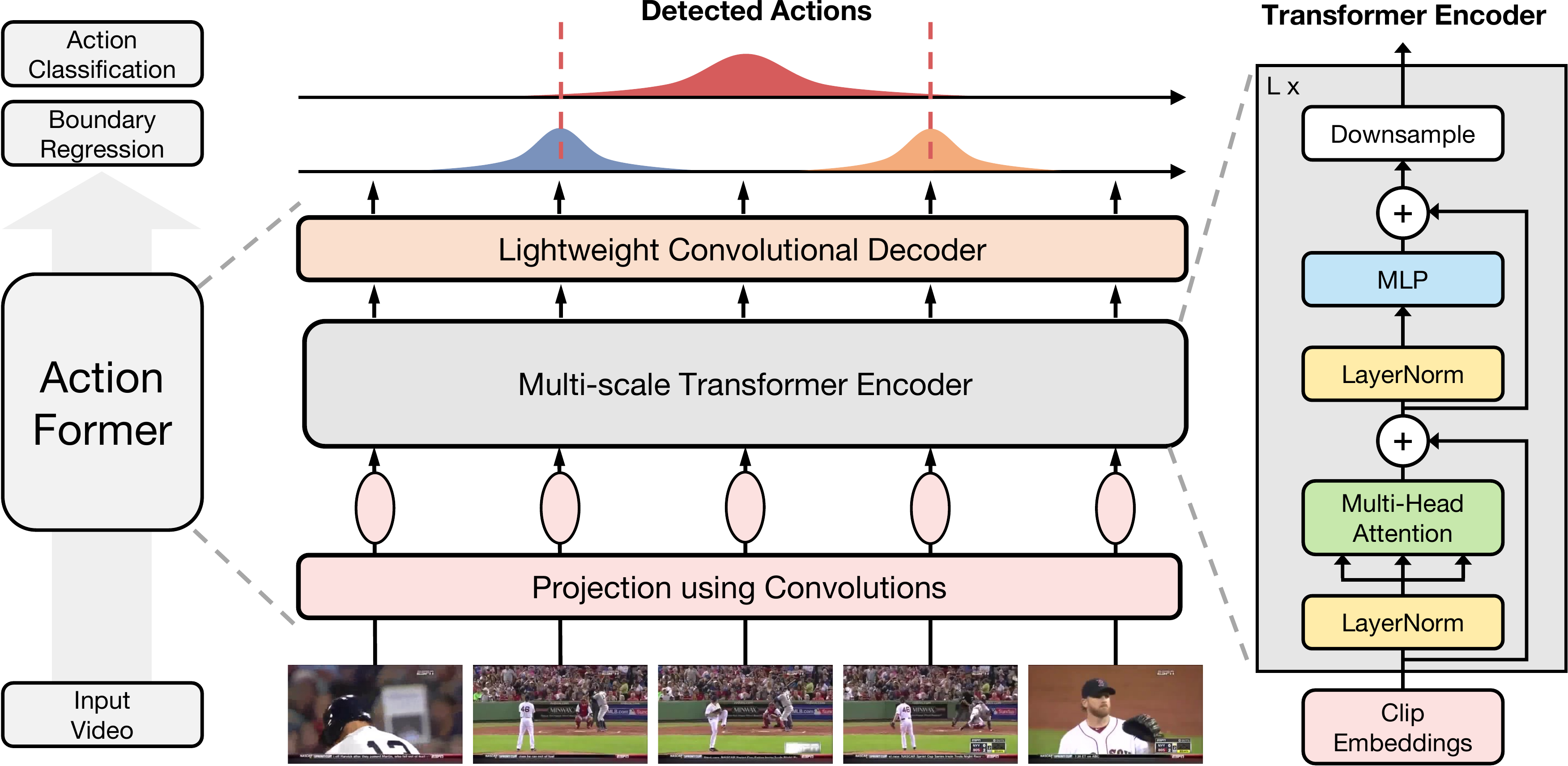}
	\caption{Overview of our ActionFormer. Our method builds a Transformer based model to detect an action instance by classifying every moment and estimating action boundaries. Specifically, ActionFormer first extracts a sequence of video clip features, and embeds each of these features. The embedded features are further encoded into a feature pyramid using a multi-scale Transformer (right). The feature pyramid is then examined by shared classification and regression heads, producing an action candidate at every time step. Our method provides a single-stage anchor-free model for temporal action localization with strong performance across several datasets.}
	\label{fig:overview}
\end{figure*}

\smallskip
\noindent \textbf{A Simple Representation for Action Localization}. Our method builds on an anchor-free representation for action localization, inspired by~\cite{a2nettip2020,afsdcvpr2021}. The key idea is to classify each moment as either one of the action categories or the background, and further regress the distance between this time step and the action's onset and offset. In doing so, we convert the structured output prediction problem ($\mathbf{X}=\{\mathbf{x}_1, \mathbf{x}_2, ..., \mathbf{x}_T\} \rightarrow \mathbf{Y} = \{\mathbf{y}_1, \mathbf{y}_2, \ldots, \mathbf{y}_N\}$) into a more approachable sequence labeling problem
\begin{equation}
\mathbf{X}=\{\mathbf{x}_1, \mathbf{x}_2, ..., \mathbf{x}_T\} \rightarrow \mathbf{\hat{Y}}=\{\mathbf{\hat{y}}_1, \mathbf{\hat{y}}_2, ..., \mathbf{\hat{y}}_T\}.
\end{equation}
The output $\mathbf{\hat{y}}_t=(p(a_t), d_t^s, d_t^e)$ at time $t$ is defined as 
\squishlist
\item $p(a_t)$ consists of $C$ values, with each representing a binomial variable indicating the probability of action category $a_t$ ($\in \{1, 2, \ldots, C\}$) at time $t$. This can be considered as the outputs of $C$ binary classification. 
\item $d_t^s>0$ and $d_t^e>0$ are the distance between the current time $t$ to the action's onset and offset, respectively. $d_t^s$ and $d_t^e$ are not defined if the time step $t$ lies on the background. 
\squishend

\smallskip
Intuitively, this formulation considers \emph{every moment} $t$ in the video $\mathbf{X}$ as an action candidate, recognizes the action's category $a_t$, and estimates the distances between current step and the action boundaries ($d_t^s$ and $d_t^e$) if an action presents. Action localization results can be readily decoded from $\mathbf{\hat{y}}_t=(p(a_t), d_t^s, d_t^e)$ by
\begin{equation}
    a_t = \argmax \ p(a_t), \quad s_t = t-d_t^s, \quad e_t=t+d_t^e.
\end{equation}%

\noindent \textbf{Method Overview}. Our model --- ActionFormer learns to label an input video sequence $f(\mathbf{X}) \rightarrow \mathbf{\hat{Y}}$. Specifically, $f$ is realized using a deep model. ActionFormer follows an encoder-decoder architecture proven successful in many vision tasks, and decomposes $f$ as $h\circ g$. Here $g\colon \mathbf{X} \rightarrow \mathbf{Z}$ encodes the input into a latent vector $\mathbf{Z}$, and $h\colon \mathbf{Z} \rightarrow \mathbf{\hat{Y}}$ subsequently decodes $\mathbf{Z}$ into the sequence label $\mathbf{\hat{Y}}$.

Fig.\ \ref{fig:overview} presents an overview of our model. Importantly, our encoder $g$ is parameterized by a Transformer network~\cite{transformernips2017}. Our decoder $h$ adopts a lightweight convolutional network. To capture actions at various temporal scales, we design a multi-scale feature representation $\mathbf{Z}=\{\mathbf{Z}^1, \mathbf{Z}^2, \ldots, \mathbf{Z}^L\}$ forming a feature pyramid with varying resolutions. Note that our model operates on a temporal axis defined by feature grids rather than the absolute time, allowing it to adapt to videos with different frame rates. We now describe the details of our model.

\subsection{Encode Videos with Transformer}\label{sec:method:encoder}
Our model first encodes an input video $\mathbf{X}=\{\mathbf{x}_1, \mathbf{x}_2, \ldots, \mathbf{x}_T\}$ into a multiscale feature representation $\mathbf{Z}=\{\mathbf{Z}^1, \mathbf{Z}^2, \ldots, \mathbf{Z}^L\}$ using an encoder $g$. The encoder $g$ consists of (1) a projection function using a convolutional network that embeds each feature ($\mathbf{x}_t$) into a $D$-dimensional space; and (2) a Transformer network that maps the embedded features to the output feature pyramid $\mathbf{Z}$.

\smallskip
\noindent \textbf{Projection}. Our projection $\mathbf{E}$ is a shallow convolutional network with ReLU as the activation function, defined as
\begin{equation}
    \mathbf{Z}^0 = [ \mathbf{E}(\mathbf{x}_1), \mathbf{E}(\mathbf{x}_2), \ldots , \mathbf{E}(\mathbf{x}_T) ]^T,
\end{equation}%
where $\mathbf{E}(\mathbf{x}_i) \in \mathbb{R}^{D}$ is the embedded feature of $\mathbf{x}_i$. Adding convolutions before a Transformer network was recently found helpful to better incorporate local context for time series data~\cite{li2019enhancing} and to stabilize the training of vision Transformers~\cite{xiao2021early}. An position embedding~\cite{transformernips2017} $\mathbf{E}_{pos}  \in \mathbb{R}^{T \times D}$ can be optionally added. However, we find that doing so will decrease the performance of the model, and have thus removed position embeddings in our model by default. 

\smallskip
\noindent \textbf{Local Self-Attention}. The Transformer network further takes $\mathbf{Z}^0$ as input. The core of a Transformer is  self-attention~\cite{transformernips2017}. We briefly introduce the key idea to make the paper self-contained. Concretely, self-attention computes a weighted average of features with the weight proportional to a similarity score between pairs of input features. Given $\mathbf{Z}^0 \in \mathbb{R}^{T \times D}$ with $T$ time steps of $D$ dimensional features, $\mathbf{Z}^0$ is projected using $\mathbf{W}_Q \in \mathbb{R}^{D \times D_q}$, $\mathbf{W}_K \in \mathbb{R}^{D \times D_k}$, and $\mathbf{W}_V \in \mathbb{R}^{D\times D_v}$ to extract feature representations $\mathbf{Q}$, $\mathbf{K}$, and $\mathbf{V}$, referred to as query, key and value respectively with $D_k = D_q$. The outputs $\mathbf{Q}$, $\mathbf{K}$, $\mathbf{V}$ are computed as
\begin{equation}
    \mathbf{Q} = \mathbf{\mathbf{Z}^0} \mathbf{W}_Q, \quad
    \mathbf{K} = \mathbf{\mathbf{Z}^0} \mathbf{W}_K, \quad
    \mathbf{V} = \mathbf{\mathbf{Z}^0} \mathbf{W}_V.
\end{equation}%
The output of self-attention is given by
\begin{equation}\label{eq:trueatten}
    \mathbf{S} = \text{softmax}\left( \mathbf{Q} \mathbf{K}^T / \sqrt{D_q} \right) \mathbf{V},
\end{equation}%
where $\mathbf{S}\in \mathbb{R}^{T \times D}$ and $\text{softmax}$ is performed {\it row-wise}. A multiheaded self-attention ($\operatorname{MSA}$) further adds several self-attention operations in parallel. 

A main advantage of $\operatorname{MSA}$ is the ability to integrate temporal context across the full sequence, yet such a benefit comes at the cost of computation. A vanilla $\operatorname{MSA}$ has a complexity of $O(T^2D + D^2T)$ in both memory and time, and is thus highly inefficient for long videos. There has been several recent work on efficient self-attention~\cite{xiong2021nystromformer,Beltagy2020Longformer,wang2020linformer,choromanski2020rethinking}. Here we adapt the local self-attention from~\cite{choromanski2020rethinking} by limiting the attention within a local window. Our intuition is the temporal context beyond a certain range is less helpful for action localization. Such a local self-attention significantly reduces the complexity to $O(W^2TD + D^2T)$ with $W$ the local window size ($\ll T$). Importantly, local self-attention is used in tandem with the multiscale feature representation $\mathbf{Z}=\{\mathbf{Z}^1, \mathbf{Z}^2, \ldots, \mathbf{Z}^L\}$, using the same window size on each pyramid level. With this design, a small window size (19) on a downsampled feature map (16x) will cover a large temporal range (304). 

\smallskip
\noindent \textbf{Multiscale Transformer}. We now present the design of our Transformer encoder. Our Transformer has $L$ Transformer layers with each layer consisting of alternating layers of local multiheaded self-attention ($\operatorname{MSA}$) and $\operatorname{MLP}$ blocks. Moreover, LayerNorm ($\operatorname{LN}$) is applied before every $\operatorname{MSA}$ or $\operatorname{MLP}$ block, and residual connection is added after every block. GELU is used for the $\operatorname{MLP}$. To capture actions at different temporal scales, a downsampling operator $\operatorname{\downarrow}(\cdot)$ is optionally attached. This is given by
\begin{equation}
\begin{split}
    \mathbf{\bar{Z}}^\ell &= \alpha^\ell \operatorname{MSA}(\operatorname{LN}(\mathbf{Z}^{\ell-1})) + \mathbf{Z}^{\ell-1}, \quad
    \mathbf{\hat{Z}}^\ell = \bar{\alpha}^\ell\operatorname{MLP}(\operatorname{LN}(\mathbf{\bar{Z}}^{\ell})) + \mathbf{\bar{Z}}^{\ell}, \\
    \mathbf{Z}^\ell &= \operatorname{\downarrow}(\mathbf{\hat{Z}}^{\ell}), \quad \ell=1\ldots L, 
\end{split}
\end{equation}%
where $\mathbf{Z}^{\ell-1}, \mathbf{\bar{Z}}^{\ell}, \mathbf{\hat{Z}}^{\ell}\in \mathbb{R}^{T^{\ell-1} \times D}$ and $\mathbf{Z}^\ell \in \mathbb{R}^{T^{\ell} \times D}$. $T^{\ell-1} / T^{\ell}$ is the downsampling ratio. $\alpha^\ell$ and $\bar{\alpha}^\ell$ are learnable per-channel scaling factors as in~\cite{touvron2021going}.

The downsampling operator $\operatorname{\downarrow}$ is implemented using a strided depthwise 1D convolution due to its efficiency. We use 2x downsampling for our model. Our Transformer block is shown in Fig.\ \ref{fig:overview} (right). Our model further combines several Transformer blocks with downsampling in between, resulting in a feature pyramid $\mathbf{Z}=\{\mathbf{Z}^1, \mathbf{Z}^2, \ldots, \mathbf{Z}^L\}$.

\subsection{Decoding Actions in Time}\label{sec:method:decoder}
Next, our model decodes the feature pyramid $\mathbf{Z}$ from the encoder $g$ into the sequence label $\mathbf{\hat{Y}}=\{\mathbf{\hat{y}}_1, \mathbf{\hat{y}}_2, \ldots, \mathbf{\hat{y}}_T\}$ using the decoder $h$. Our decoder is a lightweight convolutional network with a classification and a regression head. 

\smallskip
\noindent \textbf{Classification Head}. 
Given the feature pyramid $Z$, our classification head examines each moment $t$ across all $L$ levels on the pyramid, and predicts the probability of action $p(a_{t})$ at every moment $t$.\footnote{Without loss of clarity, we drop the index of the pyramid $\ell$.} This is realized using a lightweight 1D convolutional network attached to each pyramid level with its parameters shared across all levels. Our classification network is implemented using 3 layers of 1D convolutions with kernel size=3, layer normalization (for the first 2 layers), and ReLU activation. A sigmoid function is attached to each output dimension to predict the probability of $C$ action categories. Adding layer normalization slightly boosts the performance as we will demonstrate in our ablation. 

\smallskip
\noindent \textbf{Regression Head}. 
Similar to our classification head, our regression head examines every moment $t$ across all $L$ levels on the pyramid. The difference is that the regression head predicts the distances to the onset and offset of an action ($d^s_{t}$, $d^e_{t}$), only if the current time step $t$ lies in an action. An output regression range is pre-specified for each pyramid level. The regression head, again, is implemented using a 1D convolutional network following the same design of the classification network, except that a ReLU is attached at the end for distance estimation. 

\subsection{ActionFormer: Model Design}\label{sec:method:details}

Putting things together, ActionFormer is conceptually simple: each feature on the feature pyramid $\mathbf{Z}$ outputs an action score $p(a)$ and the corresponding temporal boundaries $(s, e)$, which are then used to decode an action candidate. Notwithstanding the simplicity, we find that several key architecture designs are important to ensure a strong performance. We discuss these design choices here. 

\smallskip
\noindent \textbf{Design of the Feature Pyramid}. A critical component of our model is the design of the temporal feature pyramid $\mathbf{Z}=\{\mathbf{Z}^1, \mathbf{Z}^2, \ldots, \mathbf{Z}^L\}$. The design choices include (1) the number of levels within the pyramid; (2) the downsampling ratio between successive feature maps; and (3) the output regression range of each pyramid level. Inspired by the design of feature pyramid in modern object detectors (FPN~\cite{fpncvpr2017} and FCOS~\cite{fcosiccv2019}), we simplify our design choices by using a 2x downsampling of the feature maps, and roughly enlarging the output regression range by 2 accordingly. We explore different design choices in our ablation.

\smallskip
\noindent \textbf{Loss Function}. Our model outputs $(p(a_t), d_t^s, d_t^e)$ for every moment $t$, including the probability of action categories $p(a_{t})$ and the distances to action boundaries ($d^s_{t}$, $d^e_{t}$). Our loss function, again following minimalist design, only has two terms: (1) $\mathcal L_{cls}$ a focal loss~\cite{focaliccv2017} for $C$ way binary classification; and (2) $\mathcal L_{reg}$ a DIoU loss~\cite{diouaaai2020} for distance regression. The loss is defined for each video $X$ as
\begin{equation}
    \mathcal L = \sum_{t}\left(
    \mathcal L_{cls} + \lambda_{reg}\mathbbm{1}_{c_t}\mathcal L_{reg}\right) / {T_{+}},
\label{eq:loss}
\end{equation}
where $T_{+}$ is the total number of positive samples. $\mathbbm{1}_{c_t}$ is an indicator function that denotes if a time step $t$ is within an action, \ie a positive sample. $\mathcal L$ is applied to all levels on the output pyramid, and averaged across all video samples during training. $\lambda_{reg}$ is a coefficient balancing the classification and regression loss. We set $\lambda_{reg}$=1 by default and study the choice of $\lambda_{reg}$ in our ablation.

Importantly, $\mathcal L_{cls}$ uses Focal loss~\cite{fcosiccv2019} to recognize $C$ action categories. Focal loss naturally handles imbalanced samples --- there are much more negative samples than positive ones. Moreover, $\mathcal L_{reg}$ adopts a differentiable IoU loss~\cite{gioucvpr2019}. $\mathcal L_{reg}$ is only enabled when the current time step contains a positive sample.

\smallskip
\noindent \textbf{Center Sampling}. During training, we find it helpful to adapt a center sampling strategy similar to~\cite{fcosiccv2019,zhang2020bridging}, as we will show in our ablation study. Specifically, when determining the positive samples, only time steps within an interval around the center of an action are considered positive, where the duration of interval is proportional to the feature stride of the current pyramid level $\ell$. More precisely, given an action centered at $c$, any time step $t \in [c - \alpha T / T^{\ell},  c + \alpha T / T^{\ell}]$ at the pyramid level $\ell$ is considered as positive, where $\alpha=1.5$. Center sampling does not impact model inference, yet encourages higher scores around action centers.

\subsection{Implementation Details}\label{sec:method:implement}

\smallskip
\noindent \textbf{Training}. Following~\cite{actiontransformercvpr2019}, we use Adam~\cite{kingma2014adam} with warm-up for training. The warm-up stage is critical for model convergence and good performance, as also pointed out by~\cite{liu2020understanding}. When training with variable length input, we fix the maximum input sequence length, pad or cropped the input sequences accordingly, and add proper masking for operations in the model. This is equal to training with sliding windows as in~\cite{a2nettip2020}. Varying the maximum input sequence length during training has little impact to the performance, as shown in our ablation. 

\smallskip
\noindent \textbf{Inference}. At inference time, we feed the full sequences into the model, as no position embeddings are used in the model. Our model takes the input video $\mathbf{X}$, and outputs $\{(p(a_{t}), d_{t}^s, d_{t}^e))\}$ for every time step $t$ across all pyramid levels. Each time step $t$ further decodes an action instance $(e_{t}=t-d_{t}^s, s_{t}=t+d_{t}^e, p(a_{t}))$. $e_t$ and $s_t$ are the onset and offset of the action, and $p(a_{t})$ is an action confidence score. The result action candidates are further processed using Soft-NMS~\cite{softnmsiccv2017} to remove highly overlapping instances, leading to the final outputs of actions.

\smallskip
\noindent \textbf{Network Architecture}. We used 2 convolutions for projection, 7 Transformer blocks for the encoder (all using local attention and with 2x downsampling for the last 5), and separate classification and regression heads as the decoder. The regression range on each pyramid level was normalized by the stride of the features. More details are presented in the appendix~\ref{sec:appendix:details}.

\section{Experiments and Results}
\label{sec:exp}
We now present our experiments and results. Our main results include benchmarks on THUMOS14~\cite{thumoscviu2017}, ActivityNet-1.3~\cite{activitynetcvpr2015} and EPIC-Kitchens 100~\cite{epicarxiv2020}. Moreover, we provide extensive ablation studies of our model.

\smallskip
\noindent \textbf{Evaluation Metric}. For all datasets, we report the standard mean average precision~(\textit{m}AP) at different temporal intersection over union~(tIoU) thresholds, widely used to evaluate TAL methods. tIoU is defined as the intersection over union between two temporal windows, \ie the 1D Jaccard index. Given a tIoU threshold, \textit{m}AP computes the mean of average prevision across all action categories. An average \textit{m}AP is also reported by averaging across several tIoUs.

\smallskip
\noindent \textbf{Baseline and Comparison}. For our main results on THUMOS14~\cite{thumoscviu2017} and ActivityNet-1.3~\cite{activitynetcvpr2015}. We compare to a strong set of baselines, including both two-stage (\eg G-TAD~\cite{gtadcvpr2020}, BC-GNN~\cite{bcgnneccv2020}, TAL-MR~\cite{bottumuptaleccv2020}) and single-stage (\eg A2Net~\cite{a2nettip2020}, GTAN~\cite{gaussiancvpr2019}, AFSD~\cite{afsdcvpr2021}, TadTR~\cite{tadtrarxiv2021}) methods for TAL. Our close competitors are those single-stage methods. Despite our best attempt for a fair comparison, we recognize some of our baselines used different setups (\eg video features). Our experiment setup follows previous works~\cite{cmcscvpr2019,bottumuptaleccv2020}. And our intention here is to compare our results to the best results previously reported.

\subsection{Results on THUMOS14}\label{sec:exp:thumos}

\smallskip
\noindent \textbf{Dataset}.
THUMOS14~\cite{thumoscviu2017} dataset contains 413 untrimmed videos with 20 categories of actions. The dataset is divided into two subsets: validation set and test set. The validation set contains 200 videos and the test set contains 213 videos. Following the common practice~\cite{gtadcvpr2020,bottumuptaleccv2020,bsneccv2018,bmniccv2019}, we use the validation set for training and report results on the test set. 

\smallskip
\noindent \textbf{Experiment Setup}.
We used two-stream I3D~\cite{kinetics2017cvpr} pretrained on Kinetics to extract the video features on THUMOS14, following~\cite{cmcscvpr2019,bottumuptaleccv2020}.
\textit{m}AP@$[0.3$:$0.1$:$0.7]$ was used to evaluate our model. 
A window size of 19 was used for local self-attention based on our ablation. Further details are described in the appendix~\ref{sec:appendix:details}. To show that our method can adapt to different video features, we also consider the pre-training method from~\cite{tspiccvworkshop2021} using an R(2+1)D network~\cite{res2p1d2018cvpr}. %

\begin{table*}[t]
 \centering
 \caption{\label{tab:result_thumosandanet} \textbf{Results on THUMOS14 and ActivityNet1.3}. We report \textit{m}AP at different tIoU thresholds. Average \textit{m}AP in $[$0.3:0.1:0.7$]$ is reported on THUMOS14 and $[$0.5:0.05:0.95$]$ on ActivityNet1.3. Best results are in \textbf{bold} and second best \underline{underlined}. Our method outperforms previous methods on THUMOS14 by a large margin, and beats previous methods when using the same features on ActivityNet1.3.}
 \resizebox{1.0\textwidth}{!}
 {
  \setlength{\tabcolsep}{2.5pt}
  \begin{tabular}{c|l|c|cccccc|cccc}
 \multirow{2}{*}{Type} & \multirow{2}{*}{Model} & \multirow{2}{*}{Feature} & \multicolumn{6}{c}{THUMOS14} & \multicolumn{4}{c}{ActivityNet1.3}\tabularnewline
 \cline{4-13}
& & & 0.3 & 0.4 & 0.5 & 0.6  & 0.7 & Avg. & 0.5 & 0.75 & 0.95 & Avg.\\

   \hline
    \multirow{17}{*}{Two-Stage} 
   & BMN~\cite{bmniccv2019} & TSN~\cite{tsn2016eccv}& 56.0 & 47.4 & 38.8 & 29.7 & 20.5 & 38.5 & 50.1 & 34.8 & 8.3 & 33.9  \tabularnewline \cline{4-13}
   & DBG~\cite{dbgaaai2020} & TSN~\cite{tsn2016eccv} & 57.8 & 49.4 & 39.8 & 30.2 & 21.7 & 39.8 & --- & --- & --- & --- \tabularnewline \cline{4-13}
   & G-TAD~\cite{gtadcvpr2020} & TSN~\cite{tsn2016eccv} & 54.5 & 47.6 & 40.3 & 30.8 & 23.4 & 39.3 & 50.4 & 34.6 & 9.0 & 34.1\tabularnewline \cline{4-13}
   & BC-GNN~\cite{bcgnneccv2020} & TSN~\cite{tsn2016eccv} & 57.1 & 49.1 & 40.4 & 31.2 & 23.1 & 40.2 & 50.6 & 34.8 & \textbf{9.4} & 34.3\tabularnewline \cline{4-13}
   & TAL-MR~\cite{bottumuptaleccv2020} & I3D~\cite{kinetics2017cvpr} & 53.9 &50.7& 45.4 &38.0& 28.5 & 43.3 & 43.5 & 33.9 & \underline{9.2} & 30.2 \tabularnewline \cline{4-13}
   & P-GCN~\cite{pgcniccv2019} & I3D~\cite{kinetics2017cvpr} & 63.6 &57.8 &49.1 & --- & --- & --- & 48.3 & 33.2 & 3.3 & 31.1 \tabularnewline \cline{4-13}
   & P-GCN~\cite{pgcniccv2019}+TSP~\cite{tspiccvworkshop2021} & R(2+1)D ~\cite{res2p1d2018cvpr} & 69.1 & 63.3 & 53.5 & 40.4 & 26.0 & 50.5 & --- & --- & --- & --- \tabularnewline \cline{4-13}
   & TSA-Net~\cite{gong2020scale} & P3D~\cite{qiu2017learning} & 61.2 & 55.9 &46.9 & 36.1 & 25.2  & 45.1 & 48.7 & 32.0 & 9.0 & 31.9 \tabularnewline \cline{4-13}
   & MUSES~\cite{liu2021multi} & I3D~\cite{kinetics2017cvpr} & 68.9 & 64.0 & 56.9 & 46.3 & 31.0 & --- & 50.0 & 35.0 & 6.6 & 34.0
     \tabularnewline \cline{4-13}
   & TCANet~\cite{tcanetcvpr2021} & TSN~\cite{tsn2016eccv} & 60.6 & 53.2 & 44.6 & 36.8 & 26.7 & 44.3 & 52.3 & 36.7 & 6.9 & 35.5 \tabularnewline \cline{4-13}
   & TCANet~\cite{tcanetcvpr2021} & SlowFast~\cite{slowfasticcv2019} & --- & --- & --- & --- & --- & --- & 54.3 & \textbf{39.1} & 8.4 & \textbf{37.6}
   \tabularnewline \cline{4-13}
   & BMN-CSA~\cite{csaiccv2021} & TSN~\cite{tsn2016eccv} & 64.4 & 58.0 & 49.2 & 38.2 & 27.8 & 47.7 & 52.4 & 36.2 & 5.2 & 35.4
     \tabularnewline \cline{4-13}
   & ContextLoc~\cite{contextlociccv2021} & I3D~\cite{kinetics2017cvpr} & 68.3 & 63.8 & 54.3 & 41.8 & 26.2 & 50.9 & \textbf{56.0} & 35.2 & 3.6 & 34.2
   \tabularnewline \cline{4-13}
   & VSGN~\cite{vsgniccv2021} & TSN~\cite{tsn2016eccv} & 66.7 & 60.4 & 52.4 & 41.0 & 30.4 & 50.2 & 52.4 & 36.0 & 8.4 & 35.1
     \tabularnewline \cline{4-13}
   & VSGN~\cite{vsgniccv2021}  & I3D~\cite{kinetics2017cvpr} & --- & --- & --- & --- & --- & --- & 52.3 & 35.2 & 8.3 & 34.7
     \tabularnewline \cline{4-13}
    & VSGN~\cite{vsgniccv2021}+TSP~\cite{tspiccvworkshop2021}  & R(2+1)D~\cite{res2p1d2018cvpr} & --- & --- & --- & --- & --- & --- & 53.3  & 36.8 & 8.1 & 35.9
    \tabularnewline \cline{4-13}
    & RTD-Net~\cite{tan2021relaxed} & I3D~\cite{kinetics2017cvpr} & 68.3 & 62.3 & 51.9 & 38.8 & 23.7 & 49.0 & 47.2  & 30.7 & 8.6 & 30.8
    \tabularnewline \cline{4-13}
     
    \hline
     \multirow{7}{*}{Single-Stage} %
     & A$^2$Net~\cite{a2nettip2020} & I3D~\cite{kinetics2017cvpr} & 58.6 &54.1& 45.5 &32.5& 17.2 & 41.6 & 43.6 & 28.7 & 3.7 & 27.8 \tabularnewline \cline{4-13}
     & GTAN~\cite{gaussiancvpr2019} & P3D~\cite{qiu2017learning} & 57.8 & 47.2 & 38.8 & --- & --- & ---  & {52.6} & 34.1 & 8.9 & 34.3 \tabularnewline \cline{4-13}
    & PBRNet~\cite{liu2020progressive} & I3D~\cite{kinetics2017cvpr} & 58.5 & 54.6 & 51.3 & 41.8 & 29.5 & --- & 54.0 & 35.0 & 9.0 & 35.0 \tabularnewline \cline{4-13}
    & AFSD~\cite{afsdcvpr2021} & I3D~\cite{kinetics2017cvpr} & 67.3 & 62.4 & 55.5 & 43.7 & 31.1 & 52.0 & 52.4 & 35.3 & 6.5 & 34.4 \tabularnewline \cline{4-13}
    & TadTR~\cite{tadtrarxiv2021} & I3D~\cite{kinetics2017cvpr} & 62.4 & 57.4 & 49.2 & 37.8 & 26.3 & 46.6 & 49.1 & 32.6 & 8.5 & 32.3 \tabularnewline \cline{4-13}
   & Ours & I3D~\cite{kinetics2017cvpr} & \textbf{82.1} & \textbf{77.8} & \textbf{71.0} & \textbf{59.4} & \textbf{43.9} & \textbf{66.8} & 53.5 & 36.2 & 8.2 & 35.6 \tabularnewline \cline{4-13}
   & Ours+TSP~\cite{tspiccvworkshop2021} & R(2+1)D~\cite{res2p1d2018cvpr} & \underline{73.4} & \underline{67.4} & \underline{59.1} & \underline{46.7} & \underline{31.5} & \underline{55.6} & \underline{54.7} & \underline{37.8} & 8.4 & \underline{36.6} \tabularnewline %
  \end{tabular}
 }
\end{table*}

\smallskip
\noindent \textbf{Results}. Table~\ref{tab:result_thumosandanet} (left) summarizes the results. Our method achieves an average \textit{m}AP of 66.8\% ($[0.3:0.1:0.7]$), with an \textit{m}AP of 71.0\% at tIoU$=$0.5 and an \textit{m}AP of 43.9\% at tIoU$=$0.7, outperforming all previous methods by a large margin (+14.1\% \textit{m}AP at tIoU$=$0.5 and +12.8\% \textit{m}AP at tIoU$=$0.7). Our results stay on top of all single-stage methods, and also beats all previous two-stage methods, including the latest ones from~\cite{bottumuptaleccv2020,tcanetcvpr2021,ctcnmm2020,csaiccv2021}. Note that our method significantly outperforms the concurrent work of TadTR~\cite{tadtrarxiv2021}, which also designed a Transformer model for TAL. With the combination of a simple design and a strong Transformer model, our method establishes new state of the art on THUMOS14, crossing the 65\% average \textit{m}AP for the first time.

\subsection{Results on ActivityNet-1.3}\label{sec:exp:anet}

\smallskip
\noindent \textbf{Dataset}.
ActivityNet-1.3~\cite{activitynetcvpr2015} is a large-scale action dataset which contains 200 activity classes and around 20,000 videos with more than 600 hours. The dataset is divided into three subsets: 10,024 videos for training, 4,926 for validation, and 5,044 for testing. Following the common practice in~\cite{bsneccv2018,bmniccv2019,gtadcvpr2020}, we train our model on the training set and report the performance on the validation set. 

\smallskip
\noindent \textbf{Experiment Setup}.
We used two-stream I3D~\cite{kinetics2017cvpr} for feature extraction.%
Following \cite{bsneccv2018,bmniccv2019,gtadcvpr2020}, the extracted features were downsampled into a fixed length of 160 using linear interpolation. For evaluation, we used \textit{m}AP@$[0.5$:$0.05$:$0.95]$ and also reported the average \textit{m}AP. %
A window size of 11 was used for local self-attention. Further implementation details can be found in the appendix~\ref{sec:appendix:details}. Moreover, we combined external classification results from~\cite{anetarxiv2017} following~\cite{bottumuptaleccv2020,gtadcvpr2020,bcgnneccv2020,pgcniccv2019}. Similarly, we consider the pre-training method from~\cite{tspiccvworkshop2021}. %

\smallskip
\noindent \textbf{Results}. Table~\ref{tab:result_thumosandanet} (right) shows the results. With I3D features, our method reaches an average \textit{m}AP of 35.6\% ($[0.5:0.05:0.95]$), outperforming all previous methods using the same features by at least 0.6\%. This boost is significant as the result is averaged across many tIoU thresholds, including those tight ones \eg 0.95. Using the pre-training method from TSP~\cite{tspiccvworkshop2021} largely improves our results (36.6\% average \textit{m}AP). Our model thus outperforms the best method with the same features~\cite{vsgniccv2021} by a major margin (+0.7\%). Again, our method outperforms TadTR~\cite{tadtrarxiv2021}. Our results are worse than TCANet~\cite{tcanetcvpr2021}---a latest two-stage method using stronger SlowFast features~\cite{slowfasticcv2019} that are not publicly available. We conjecture our method will also benefit from better features. Nonetheless, our model clearly demonstrates state-of-the-art results on this challenging dataset. 

\subsection{Results on EPIC-Kitchens 100}\label{sec:exp:epic}
\begin{table}[t]
    \centering
    \caption{\textbf{Results on EPIC-Kitchens 100 validation set}. We report \textit{m}AP at different tIoU thresholds and the average \textit{m}AP in $[0.1$:$0.1$:$0.5]$. All methods used the same SlowFast features. Our method outperforms all baselines by a large margin.}
	\label{table:all epic-kitchens results}
    {
    \scriptsize
	\setlength{\tabcolsep}{2.5pt}
	\begin{tabular}{l|l|c|c|c|c|c|c}
	{Task} & {Method} & 0.1 &0.2 & 0.3 & 0.4 & 0.5 & Avg\\
		\hline
	\multirow{3}{*}{Verb} & BMN~\cite{bmniccv2019,epicarxiv2020} & 10.8	&9.8&8.4&7.1&5.6&8.4 \\
	& G-TAD~\cite{gtadcvpr2020} & 12.1 & 11.0 & 9.4 & 8.1 & 6.5 & 9.4\\
	& Ours & \textbf{26.6} & \textbf{25.4} &\textbf{24.2}&\textbf{22.3} &\textbf{19.1} &\textbf{23.5} \\
	\hline 
	\multirow{3}{*}{Noun} & BMN~\cite{bmniccv2019,epicarxiv2020} & 10.3	&8.3&6.2&4.5&3.4&6.5 \\
	& G-TAD~\cite{gtadcvpr2020} & 11.0 & 10.0 & 8.6 & 7.0 & 5.4 & 8.4\\
	& Ours & \textbf{25.2} & \textbf{24.1} &\textbf{22.7}&\textbf{20.5} &\textbf{17.0} &\textbf{21.9} \\
	\end{tabular}}

\end{table}

\smallskip
\noindent \textbf{Dataset}.
EPIC-Kitchens 100 is the largest egocentric action dataset. The dataset contains 100 hours of videos from 700 sessions capturing cooking activities in different kitchens. In comparison to ActivityNet-1.3, EPIC-Kitchens 100 has less number of videos, yet many more instances per video (average 128 vs.\ 1.5 on ActivityNet-1.3). In comparison to THUMOS14, EPIC-Kitchens is 3 times larger in terms of video hours and more than 10 times larger in terms of action instances. These egocentric videos also include significant camera motion. This dataset thus poses new challenges for TAL. 

\smallskip
\noindent \textbf{Experiment Setup}.
We used a SlowFast network~\cite{slowfasticcv2019} pre-trained on EPIC-Kitchens for feature extraction. This model is provided by~\cite{epicarxiv2020}. %
Our model was trained on the training set and evaluated on the validation set. A window size of 9 was used for local self-attention. For evaluation, we used \textit{m}AP@$[0.1$:$0.1$:$0.5]$ and report the average \textit{m}AP following~\cite{epicarxiv2020}. In this dataset, an action is defined as a combination of a verb (action) and a noun (object). As this dataset was recently released, we are only able to compare our methods to BMN~\cite{bmniccv2019} and G-TAD~\cite{gtadcvpr2020}, both using the same SlowFast features provided by~\cite{epicarxiv2020}. Again, implementation details are described in the appendix~\ref{sec:appendix:details}.

\smallskip
\noindent \textbf{Results}.
Table~\ref{table:all epic-kitchens results} presents the results. Our method achieves an average \textit{m}AP ($[0.1$:$0.1$:$0.5]$) of 23.5\% and 21.9\% for verb and noun, respectively. Our results again largely outperform the strong baselines of BMN~\cite{bmniccv2019} and G-TAD~\cite{gtadcvpr2020} by over 13.5\% in absolute percentage points. An interesting observation is that the gaps between our results and BMN / G-TAD are much larger on EPIC-Kitchens 100. A possible reason is that ActivityNet has a small number of actions per video (1.5), leading to imbalanced classification for our model; only a few moments (around the action center) are labeled positive while all rest are negative.

We adapt ActionFormer for EPIC-Kitchens 100 2022 Action Detection challenge. By combining features from SlowFast~\cite{slowfasticcv2019} and ViViT~\cite{arnab2021vivit}, ActionFormer achieves 21.36\% / 20.95\% average \textit{m}AP for actions on the validation / test set. Our results ranked 2nd with a gap of 0.32 average \textit{m}AP to the top solution.

\subsection{Ablation Experiments}\label{sec:exp:ablation}
We conduct extensive ablations on THUMOS14 to understand our model design. Results are reported using I3D features with a fixed random seed for training. Further ablations on loss weight, maximum input length during training, input temporal feature resolution and error analysis can be found in the appendix~\ref{sec:appendix:ablation}.

\smallskip
\noindent \textbf{Baseline: A Convolutional Network}. Our ablation starts by re-implementing a baseline anchor-free method (AF Base) as described in~\cite{a2nettip2020,afsdcvpr2021} (Table~\ref{tab-ablations}a row 1-2). This baseline shares the same action representation as our model, yet uses a 1D convolutional network as the encoder. We roughly match the number of layers and parameters of this baseline to our model. See the appendix~\ref{sec:appendix:details} for more details. This baseline achieves an average \textit{m}AP of 46.6\% on THUMOS14 (Table~\ref{tab-ablations}a row 3), outperforms the numbers reported in~\cite{afsdcvpr2021} by 6.2\%. We attribute the difference to variations in architectures and training schemes. This baseline, when using score fusion, reaches 52.9\% average \textit{m}AP (Table~\ref{tab-ablations}a row 4). 

\smallskip
\noindent \textbf{Transformer Network}. Our next step is to simply replace the 1D convolutional network with our Transformer model using vanilla self-attention. This model achieves an average \textit{m}AP of 62.7\% (Table~\ref{tab-ablations}a row 5) --- a major boost of 16.1\%. We note that this model already outperforms the best reported results (56.9\% \textit{m}AP at tIoU$=$0.5 from~\cite{liu2021multi}). This result shows that our Transformer model is very powerful for TAL, and serves as the main course of performance gain.

\begin{table}[t]\centering
\caption{\textbf{Ablation} studies on THUMOS14. We report \textit{m}AP at tIoU$=$0.5 and 0.7, and the average \textit{m}AP in $[0.3:0.1:0.7]$. Results are without score fusion unless specified.\label{tab-ablations}} 
\subfloat[\textbf{Model Design}: We start from a baseline 1D convolutional network (AF Base), and gradually replace convolutions with our Transformer model, add layer norm to heads (LN), enable center sampling during training (CTR), explore position encoding (PE), and fuse classification scores (Fusion).]
{
  \scriptsize
  \setlength{\tabcolsep}{2.5pt}
  \begin{tabular}{l|c|cccc|c|c|c}
  Method             & Backbone & LN & CTR & PE & Fusion &0.5  & 0.7  & Avg  \\
  \hline
  AF Base~\cite{a2nettip2020}   & Conv     &    &     &                   &     & 36.6 & 15.0 & 34.2 \\
  AF Base~\cite{afsdcvpr2021}   & Conv     &    &     &                   &     & 31.0 & 19.0 & 40.4 \\
  AF Base (Our Impl) & Conv     &    &     &                              &     & 48.0 & 29.4 & 46.6 \\
  AF Base (Our Impl) & Conv     &    &     &                     & \checkmark   & 54.6 & 33.4 & 52.9\\\hline
  Ours               & Trans    &    &     &                              &     & 66.8 & 38.4 & 62.7 \\
  Ours               & Trans    & \checkmark  &     &                     &     & 69.0 & 43.0 & 65.4 \\
  Ours               & Trans    & \checkmark  & \checkmark   & \checkmark           &     & 70.4 & 43.6 & 66.7 \\
  Ours               & Trans    & \checkmark  & \checkmark   &  & \checkmark    & 66.0 & 41.7 & 62.1 \\
  Ours               & Trans    & \checkmark  & \checkmark   &  &    & \textbf{71.0} & \textbf{43.9} & \textbf{66.8} \\
  Ours (win size=19)  & Trans    & \checkmark  & \checkmark   &  &    & \textbf{71.0} & \textbf{43.9} & \textbf{66.8}
  \end{tabular}
}
\\
\resizebox{0.49\textwidth}{!}
{
\subfloat[\textbf{Local Window Size}: We additionally report MACs and normalized run time by varying the local window size for self-attention in our model, using an input of $2304$ time steps (5 minutes on THUMOS14). Time is normalized by setting AF Base to 1.0x.]
{
  \setlength{\tabcolsep}{2.5pt}
  \begin{tabular}{l|c|c|c|c|c|c}
  Method  & Win Size & 0.5           & 0.7           & Avg           & GMACs  & Time\\
  \hline
  AF Base & N/A      & 48.0          & 29.4          & 46.6          &  45.6   & 1.0x  \\
  \hline
  Ours    & 9        & 70.5          & 42.7          & 66.5         &  45.2   & 2.0x  \\
  Ours    & 19       & \textbf{71.0}          & \textbf{43.9}          & \textbf{66.8}          & 45.3 & 2.0x     \\
  Ours    & 25       & 70.3          & 43.9          & 66.4          & 45.4   & 2.0x    \\
  Ours    & 37       & \textbf{71.0}          & 43.1          & 66.7         & 45.5 & 2.0x      \\
  Ours    & Full     & \textbf{71.0}          & \textbf{43.9}          & \textbf{66.8}          & 57.8  & 2.2x  
  \end{tabular}
}
}~
\resizebox{0.49\textwidth}{!}{
\subfloat[\textbf{Design of Feature Pyramid}: We vary (1) the number of pyramid levels and (2) the initial regression range, and report \textit{m}AP and the average \textit{m}AP.]
{
  \setlength{\tabcolsep}{2.5pt}
  \begin{tabular}{l|c|c|c|c|c}
  Method  & \# Levels & Init Range & 0.5           & 0.7           & Avg          \\
  \hline
  Ours    & 1     & [0, $+\infty$) & 51.8  & 15.8  & 47.6    \\
  \hline
  Ours    & 3     & [0, 4)  & 64.4  & 31.5  & 60.1    \\
  Ours    & 3     & [0, 8)  & 61.4  & 30.0  & 57.6    \\
  Ours    & 3     & [0, 16) & 54.2  & 19.2  & 50.2    \\
  \hline
  Ours    & 4     & [0, 4) & 67.4  & 39.7  & 63.7    \\
  Ours    & 5     & [0, 4) & 70.2  & 42.2  & 65.5    \\
  Ours    & 6     & [0, 4) & \textbf{71.0}  & \textbf{43.9}  & \textbf{66.8}    \\
  Ours    & 7     & [0, 4) & 70.6  & 43.2  & 66.2    \\
  \end{tabular}
}
}
\end{table}

\smallskip
\noindent \textbf{Layer Norm, Center Sampling, Position Encoding, \& Score Fusion}. We further add layer norm in the classification and regression heads, apply center sampling during training, and explore position encoding as well as score fusion (Table~\ref{tab-ablations}a row 6-9). Adding layer norm boosts the average \textit{m}AP by 2.7\%, and using center sampling further improves the performance by 1.4\%. The commonly used position encoding, however, does not bring performance gain. We postulate that our projection using convolutions as well as the depthwise convolutions in our Transformer blocks already leak the location information, as also pointed out in~\cite{xie2021segformer}. Further fusing the classification scores will decrease the largely performance. As a reference, when replacing the vanilla self-attention with a local version (window size=19), the average \textit{m}AP remains the same.

\smallskip
\noindent \textbf{Window Size for Local Self-Attention}. Next, we study the effects of window size for local self-attention in our model. We vary the window size, re-train the model, and present both model accuracy, complexity (in GMACs), and run time in Table~\ref{tab-ablations}b. All results are reported without score fusion. Due to our design of a multiscale feature pyramid, even using the global self-attention only leads to 26\% increase in MACs when compared to the baseline convolutional network. Reducing the window size cuts down the MACs yet maintains a similar accuracy. In addition to MACs, we also evaluate the normalized run time of these models on GPUs, where the base convolutional model is set to 1.0x. In spite of similar MACs, Transformer-based models are roughly 2x slower in run time compared to a convolution-based model (AF Base). It is known that self-attention is not easily parallelizable on GPUs. Also, our current implementation of Transformer used PyTorch primitives~\cite{paszke2019pytorch} without leveraging customized CUDA kernels.

\smallskip
\noindent \textbf{Feature Pyramid}. Further, we study the design of the feature pyramid. As discussed in Sec.\ \ref{sec:method:details}, our design space is specified by (1) the number of pyramid levels and (2) an initial regression range for the first pyramid level. We vary these parameters and report results in Table~\ref{tab-ablations}c. We follow our best design with a local window size=19 and layer norm, and center sampling. 

First, we disable the feature pyramid and attach the heads to the feature map with the highest resolution. This is done by setting the number of pyramid to 1 with an initial regression range of [0, $+\infty$), Removing the feature pyramid results in a major performance drop (-19.3\% in average mAP), suggesting that using feature pyramid is critical for our model. Next, we set the number of pyramid levels to 3 and experiment with different initial regression ranges. The best results are achieved with the range of [0, 4). Further increase of the range decreases the mAP scores. Finally, we fix the initial regression range to [0, 4) and increase the number of pyramid levels. The performance of our method generally increases with more pyramid levels, yet is saturated when using 6 levels.

\begin{figure}[t]
\centering
\includegraphics[width=0.95\linewidth]{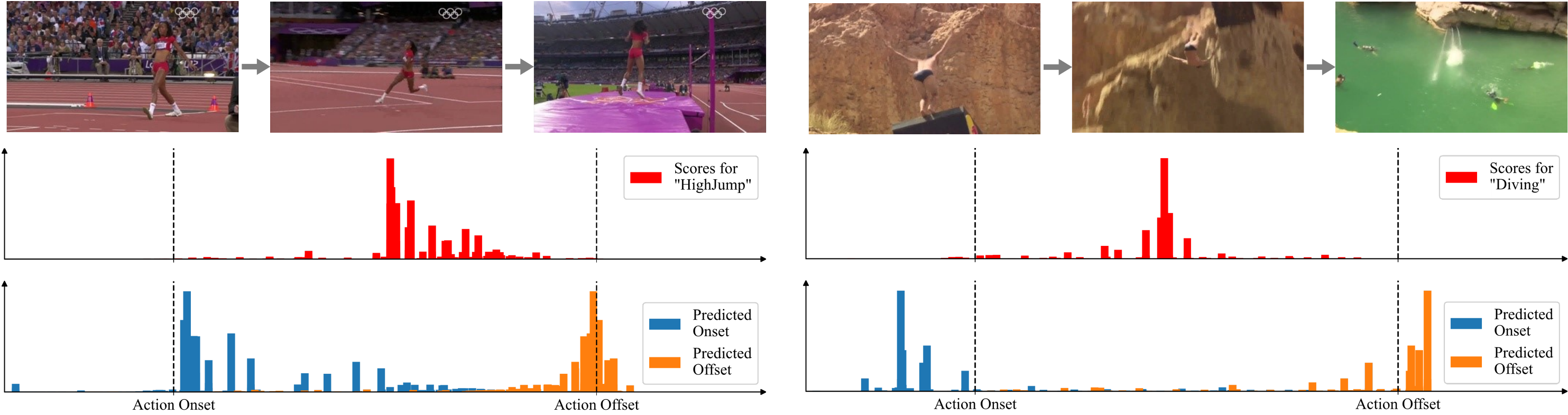}
\caption{Visualization of our results. From \emph{top} to \emph{bottom}: (1) input video frames;  (2) action scores at each time step; (3) histogram of action onsets and offsets computed by weighting the regression outputs using action scores. See more in the appendix~\ref{sec:appendix:vis}}
\label{fig:visualization}
\end{figure}

\smallskip
\noindent \textbf{Result Visualization}. Finally, we visualize the outputs of our model (before Soft-NMS) in Fig.\ \ref{fig:visualization}, including the action scores, and the regression outputs weighted by the action scores (as a weighted histogram). Our model outputs a strong peak near the center of an action, potentially due to the employment of center sampling during training. The regression of action boundaries seems less accurate. We conjecture that our regression heads can be further improved.

\section{Conclusion and Discussion}
In this paper, we presented ActionFormer---a Transformer-based method for temporal action localization. ActionFormer has a simple design, falls into the category of single-stage anchor-free method, yet achieves impressive results across several major TAL benchmarks including THUMOS14, ActivityNet-1.3, and the more recent EPIC-Kitchens 100 (egocentric videos). Through our experiments, we showed that the power of ActionFormer lies in our orchestrated design, in particular the combination of local self-attention and a multiscale feature representation to model longer range temporal context in videos. We hope that our model, notwithstanding its simplicity, can shed light on the task of temporal action localization, as well as the broader field of video understanding.

\newpage

\section{Appendix}
\appendix

\setcounter{figure}{0}
\setcounter{table}{0}
\setcounter{section}{0}
\renewcommand{\theequation}{\Alph{equation}}
\renewcommand{\thefigure}{\Alph{figure}}
\renewcommand{\thesection}{\Alph{section}}
\renewcommand{\thetable}{\Alph{table}}

\renewcommand{\tableautorefname}{Table}
\renewcommand{\figureautorefname}{Fig.}
\renewcommand{\sectionautorefname}{Sec.}
\renewcommand{\subsectionautorefname}{Sec.}
\renewcommand{\equationautorefname}{Eq.}

In the appendix, we describe (1) additional ablation experiments (\autoref{sec:appendix:ablation}); (2) further error analysis of our results (\autoref{sec:appendix:error_analysis}); (3) implementation details and how to reproduce our results (\autoref{sec:appendix:details}); (4) additional visualizations of our results (\autoref{sec:appendix:vis}); and (5) limitation of our approach and furture directions (\autoref{sec:appendix:limit}). For sections, figures, tables, and equations, we use numbers (\eg, Sec.\ 1) to refer to the main paper and capital letters (\eg, Sec.\ A) to refer to this appendix. %

\section{Additional Ablation Experiments}\label{sec:appendix:ablation}
Here we present additional ablation experiments, as mentioned in Sec.\ 4.4 of the main paper. These are omitted from the main paper due to lack of space. All experiments are reported on THUMOS14, consistent with our ablation experiments in the main paper. We follow our best design and use a local window size=19 with layer norm, center sampling, and score fusion enabled.

\medskip
\noindent \textbf{Loss Weight}.
We provide additional ablation on the loss weight $\lambda_{reg}$ in Eq.\ 7. Specifically, we varied the loss weight $\lambda_{reg} \in [0.2, 0.5, 1, 2, 5]$, retrained the model, and reported the mAP scores. The results are presented in Table~\ref{ablation3}. For a large range of $\lambda_{reg}$, our model has quite stable results with a maximum gap of 1.4\% in average mAP. $\lambda_{reg}=1$ yields the best results, as we used in all our experiments. 

\begin{table}[th!]
\caption{\label{ablation3}\textbf{Ablation study on loss weight}. We report \textit{m}AP at tIoU$=$0.5 and 0.7, and the average \textit{m}AP in $[0.3:0.1:0.7]$ on THUMOS14 by varying the loss weight $\lambda_{reg}$ in Eq.\ 7.}

\centering 
{
  \setlength{\tabcolsep}{2.5pt}
  \begin{tabular}{l|c|c|c|c}
  Method  & $\lambda_{reg}$ & 0.5           & 0.7           & Avg          \\
  \hline
  Ours    & 0.2   & 69.7          & 40.9           & 65.6          \\
  Ours    & 0.5   & \textbf{71.3}           & 42.4          & 66.7          \\
  Ours    & 1     & 71.0          & \textbf{43.9}          & \textbf{66.8}          \\
  Ours    & 2     & 69.5           & \textbf{43.9}           & 66.2         \\
  Ours    & 5     & 69.1  & 43.0  & 65.4  \\
  \end{tabular}
}
\end{table}

\noindent \textbf{Maximum Input Sequence Length during Training}. A possible explanation of our superior results is that our model might benefit from training using a long sequence (2304 time steps as in our previous experiments). Here we examine the effects of maximum input sequence length during training. Table~\ref{ablation4} reports mAP scores for different training sequence lengths. The results of our model remain fairly consistent even with much shorter input sequence length. Note that when truncating an input sequence, our training scheme is equal to training with sliding windows as in~\cite{a2nettip2020}. The differences are (1) the windows are dynamically sampled rather than pre-generated; (2) windows without foreground actions are removed. When using a input sequence length of 512, similar to what was considered in~\cite{a2nettip2020} (512), our method only has a minor drop in average mAP (-1.1\%) and significantly outperforms~\cite{a2nettip2020}.

\begin{table}[th!]
\caption{\label{ablation4}\textbf{Ablation study on maximum input sequence length during training}. We report \textit{m}AP at tIoU$=$0.5 and 0.7, and the average \textit{m}AP in $[0.3:0.1:0.7]$ on THUMOS14 by varying the maximum input length $T_{max}$ for training.}
\centering
{
  \setlength{\tabcolsep}{2.5pt}
  \begin{tabular}{l|c|c|c|c}
  Method  & $T_{max}$ & 0.5           & 0.7           & Avg          \\
  \hline
  Ours    & 576     &  69.6     & 42.5      &  65.7    \\
  Ours    & 1152     & \textbf{71.0}   & 42.7  & 66.3     \\
  Ours    & 2304     & \textbf{71.0}   & \textbf{43.9} & \textbf{66.8}  \\
  \end{tabular}
}
\end{table}

\noindent \textbf{Temporal Feature Resolution}. Some of the previous works considered video features with lower temporal resolution. For example, a feature stride of 8 was used by PGCN~\cite{pgcniccv2019} and ContextLoc~\cite{contextlociccv2021}. To understand the effects of temporal feature resolution, we downsample our input I3D features and study the performance variation when using different feature strides. Table~\ref{ablation5} report the results. When using a lower resolution (stride=8), the results of our model only drop slightly (-0.5\% in average mAP). Further reducing the resolution (\eg, stride=16) leads to larger performance degradation, yet our results remains favourable. 

\begin{table}[th!]
\caption{\label{ablation5}\textbf{Ablation study on temporal feature resolution}. We report \textit{m}AP at tIoU$=$0.5 and 0.7, and the average \textit{m}AP in $[0.3:0.1:0.7]$ on THUMOS14 by varying the feature stride.}
\centering
{
\setlength{\tabcolsep}{2.5pt}
  \begin{tabular}{l|c|c|c|c}
  Method  & stride & 0.5           & 0.7           & Avg          \\
  \hline
  Ours    & 4     &  \textbf{71.0}  & \textbf{43.9}  & \textbf{66.8}    \\
  Ours    & 8     & 69.8   & \textbf{43.9}  & 66.3    \\
  Ours    & 16    & 65.8   & 38.4  & 61.9  \\
  \end{tabular}
}
\end{table}

\section{Further Error Analyses}\label{sec:appendix:error_analysis}
We present further analyses of our results on THUMOS14 using the tool provided by~\cite{detadeccv2018}. We refer the readers to~\cite{detadeccv2018} for more details.\medskip

\begin{figure*}[th!]
    \centering
      \includegraphics[width=0.6\linewidth]{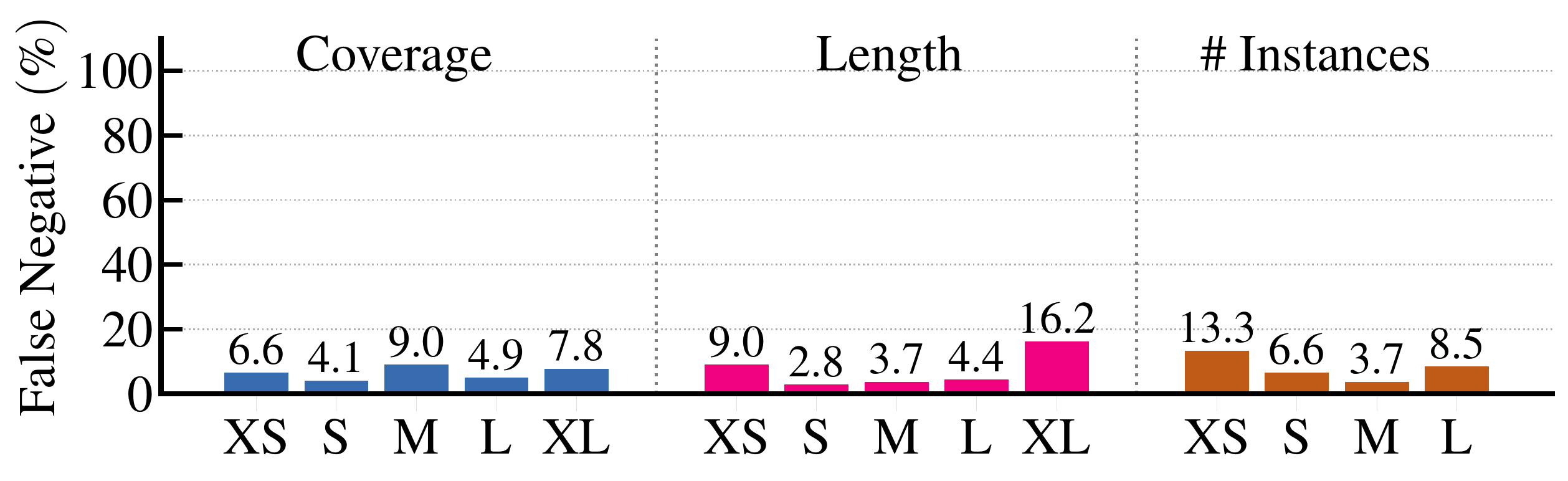}
      \caption{False negative (FN) profiling of our results on THUMOS14 using~\cite{detadeccv2018}. This figure shows the FN rates under different video contents. From this figure we can find that our model will suffer from extra short or extra long instances. Also, our model will suffer from video inputs which have a large number of action instances.}\label{fig:false negative}
\end{figure*}
\begin{figure*}[th!]
    \centering
    \includegraphics[width=0.75\linewidth]{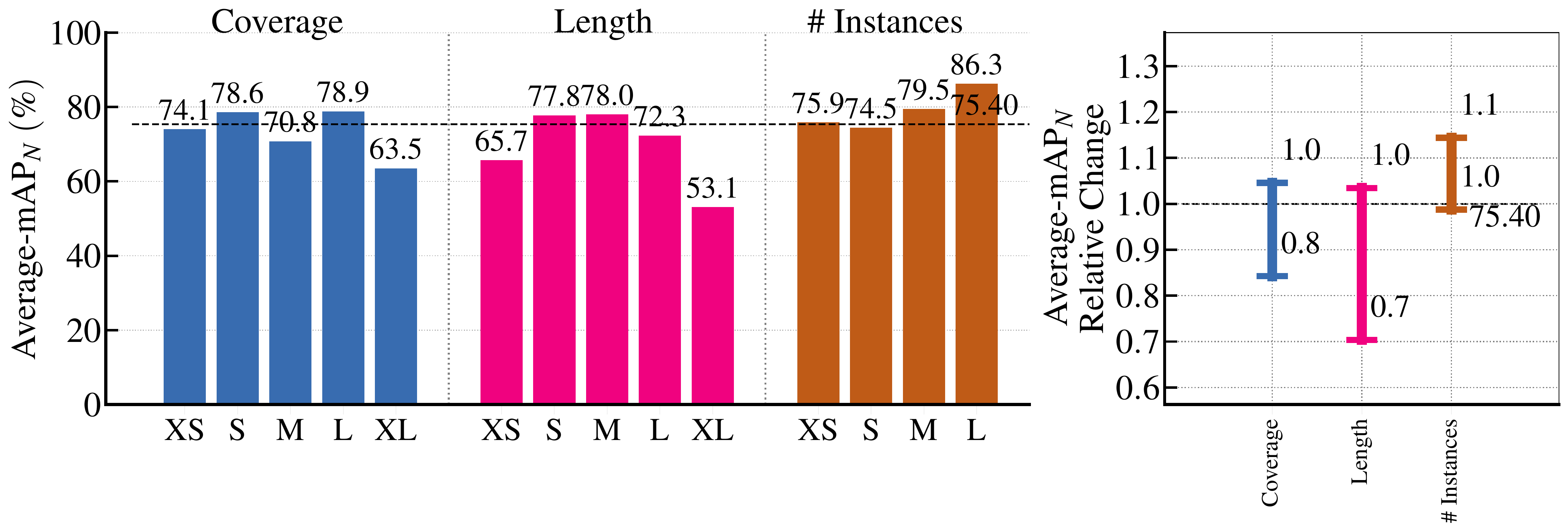}
    \caption{Sensitive analysis of our results on THUMOS14 using~\cite{detadeccv2018}. \emph{Left}: normalized mAP at tIoU$=$0.5 under different video contents. \emph{Right}: The relative normalized mAP change at tIoU$=$0.5 with respect to different characteristics of the ground truth instances.}
    \label{fig:sensitive analysis}
\end{figure*}
\begin{figure*}[th!]
    \centering
    \includegraphics[width=0.9\linewidth]{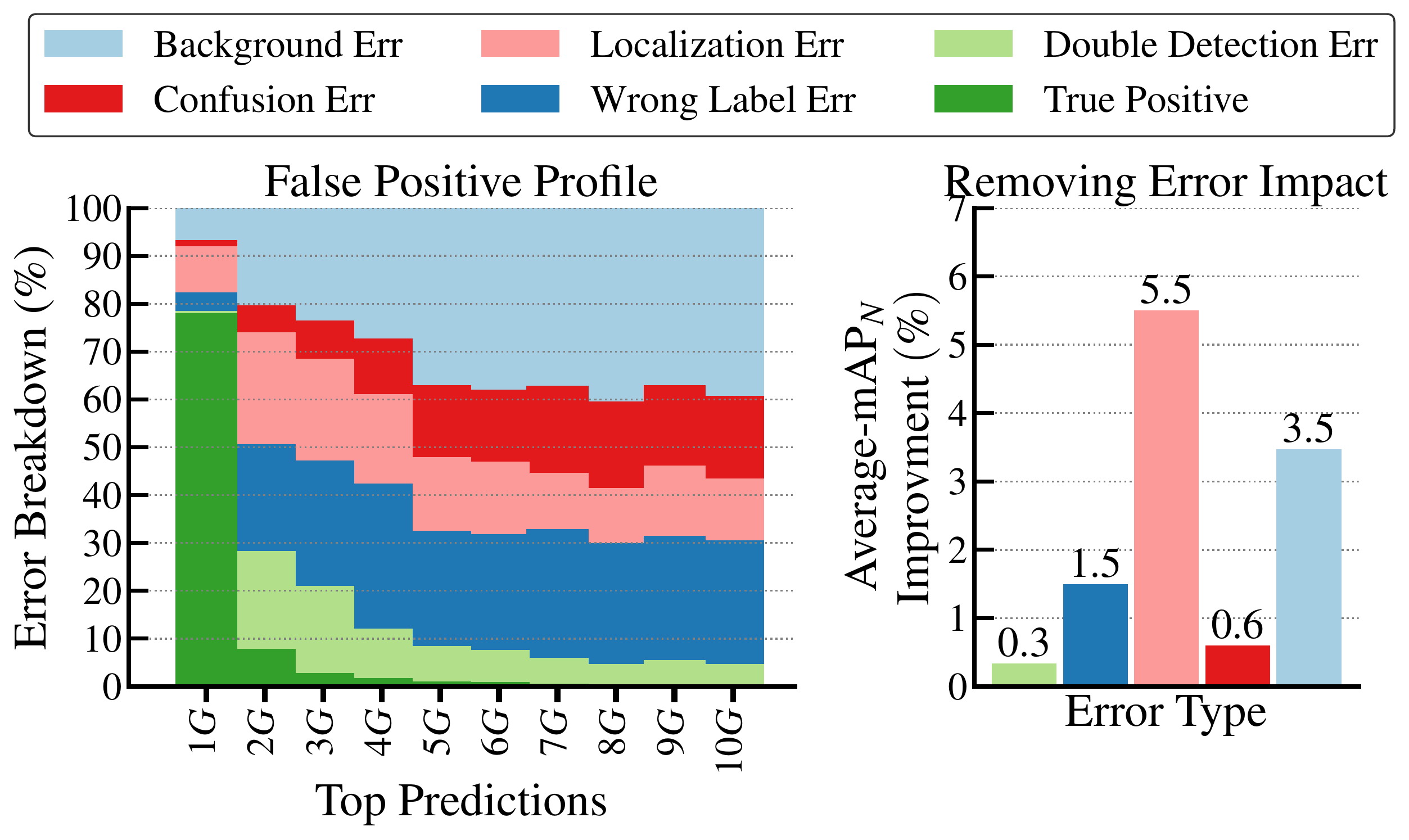}
    \caption{False positive (FP) profiling of our results on THUMOS14 using~\cite{detadeccv2018}. \emph{Left}: FP error breakdown when considering the predictions for the top-10 ground-truth (G) instances. \emph{Right}: The impact of error types. Localization error and background confusion are the top two error types.}
    \label{fig:false positive}
\end{figure*}

\noindent \textbf{Metrics}. In~\cite{detadeccv2018}, several characteristic metrics were defined given a dataset (\eg THUMOS14), including coverage, length, and the number of instances. Specifically, coverage presents the relative length of the actions (compared to the whole video), categorized into five bins: Extra Small (XS: (0, 0.02]), Small (S: (0.02, 0.04]), Medium (M: (0.04, 0.06]), Large (L:(0.06, 0.08]), and Extra Large (XL: (0.08, 1.0]). Length denotes the absolute length~(in seconds) of actions, organized into five length groups: Extra Small (XS: (0, 3]), Small (S: (3, 6]), Medium (M: (6, 12]), Long (L: (12, 18]), and Extra Long (XL: $>$ 18). Moreover, number of instances refers to the total count of instances (from the same class) in a video. This number is further divided into four parts, including Extra Small (XS: 1); Small (S: [2, 40]); Medium (M: [40, 80]); Large (L: $>$ 80).\medskip 

\noindent \textbf{Results and Analyses}. Fig.\ \ref{fig:false negative} presents the false negative profiling. In Fig.\ \ref{fig:false negative}, we breakdown the false negative rates under the different coverage, length, and the number of instances. Our results have similar false negative rates across different coverage categories, yet have much higher false negative rates on action instances that are either very shot or very long (length), and on videos that contains many action instances (\#instances). These action instances and videos are naturally more challenging. 

Fig.\ \ref{fig:sensitive analysis} presents the sensitivity analysis of our results, \ie, normalized mAP at tIoU$=$0.5 under different characteristic metrics (left) and the variance of mAP across categories (right). Our model performs better on simple context scenarios, including XS/S/M/L coverage, S/M length and XS \#instances, and worse on more complicated scenarios. The trend is similar to the false negative profiling in Fig.\ \ref{fig:false negative}. Moreover, our model is robust across different categories in coverage, length and \#instances with small variances. 

\section{Implementation Details}\label{sec:appendix:details}
We now present implementation details including the network architecture, training and inference. Further details can be found in our code.

\begin{table*}[t!]
	\centering
	\caption{\textbf{The architecture of our model.} Our network consists of (1) a Transformer encoder (first row block) and (2) a lightweight convolutional decoder with the classification / regression heads (last row block). For each layer, we list the layer name, layer parameters, the input to the layer, and the output feature size. We also include its regression range (in seconds for THUMOS14 and EPIC-Kitchens 100 and in number of time steps for ActivityNet-1.3). For convolutional layers, $k$ is the kernel size of 1D convolutions and $s$ is the stride, and $c_{i},c_{o}$ is the input and output feature channel, respectively. For Transformer Unit, $ds$ is the downsampling ratio. $T$ is the temporal length of input sequence and $D$ is the input feature dimension. For classification head, the output dimension is the number of action categories. For regression head, the output dimension is 2, \ie, distances to action onset and offset.}
    \resizebox{1.0\textwidth}{!}{
	\begin{tabular}{c|c|c|c|c|c}
	    &
		\multicolumn{1}{c|}{Name} & 
		\multicolumn{1}{c|}{Layer} & 
		\multicolumn{1}{c|}{Input} & \multicolumn{1}{c|}{\begin{tabular}[c]{@{}c@{}}Output Size\\ (T $\times$ D)\end{tabular}} & 
		\multicolumn{1}{c}{\begin{tabular}[c]{@{}c@{}}Regression\\ Range\end{tabular}} \\
		\hline
		\multirow{13}{*}{encoder} & input clip        &  -    & - & T$\times$D    & - \\
		&projection1       &  conv $k$=3, $s$=1 ($c_{i}$ = D, $c_{o}$ = 512)  & input clip & T $\times$ 512 &-\\
		&projection2       &  conv $k$=3, $s$=1 ($c_{i}$ = 512, $c_{o}$ = 512)  & projection1 & T $\times$ 512 &-\\
		&transformer0 &  Transformer Unit, $ds$=1  & projection2 &  T $\times$ 512 & -\\
		&transformer1 &  Transformer Unit, $ds$=1  & transformer0 &  T $\times$ 512 & [0, 4)\\
		&transformer2 &  Transformer Unit, $ds$=2  & transformer1 &  T/2 $\times$512 & [4, 8) \\ 
        &transformer3 & Transformer Unit, $ds$=2   & transformer2 &  T/4 $\times$512& [8, 16)\\
        &transformer4 & Transformer Unit, $ds$=2   & transformer3 &  T/8 $\times$512& [16, 32)\\
		&transformer5 & Transformer Unit, $ds$=2   & transformer4 &  T/16 $\times$512 & [32, 64)\\
		&transformer6 & Transformer Unit, $ds$=2   & transformer5 &  T/32 $\times$512 & [64, $+\infty$)\\
		\hline
		\multirow{3}{*}{\begin{tabular}[c]{@{}c@{}}decoder\\ (heads)\end{tabular}} &\multirow{3}{*}{cls / reg nets} &  conv $k$=3, $s$=1 ($c_{i}$ = 512, $c_{o}$ = 512) &  transformer1,...,transformer6 &[T/32$\times$512,$\ldots$, T$\times$512]&-\\
		& &  conv $k$=3, $s$=1 ($c_{i}$ = 512, $c_{o}$ = 512) & transformer1,...,transformer6  & [T/32$\times$512,$\ldots$, T$\times$512]&-\\
		& &  conv $k$=3, $s$=1 ($c_{i}$ = 512, $c_{o}$ = output) & transformer1,...,transformer6 & [T/32$\times$output,$\ldots$, T$\times$output]&-\\
	\end{tabular}}
	\label{table:network structure on THUMOS14 and EPIC-Kitchens}
\end{table*}

\medskip
\noindent \textbf{Network Architecture}.
We present our network architecture in Table~\ref{table:network structure on THUMOS14 and EPIC-Kitchens}, as described in Sec.\ 3. In the ablation study (Sec.\ 4), we also considered a baseline that replaces the Transformer Units in Table~\ref{table:network structure on THUMOS14 and EPIC-Kitchens} with convolution blocks, following the design of a bottleneck block in ResNet~\cite{resnetcvpr2016}. Specifically, a stack of three 1D convolutional layers were used. The kernel size of three convolutional layers were 1, 3 and 1, respectively. The expansion factor of the bottleneck block was 2. We added an extra strided convolutional layer with kernel size=1 and stride=2 to perform downsampling when necessary. 

\medskip
\noindent \textbf{Training Details}. For training, we considered both fixed length inputs (ActivityNet) and variable length inputs (THUMOS14, ActivityNet, and EPIC-Kitchens 100). For variable length inputs, we capped the input length to 2304 (around 5 minutes on THUMOS14 and around 20 minutes on EPIC-Kitchens 100), and randomly selected a subset of consecutive clips from an input video. Position embedding was disabled by default except for ActivityNet. Model EMA~\cite{huang2017snapshot} and gradient clipping were also implemented to further stabilize the training. Hyperparameters were slightly different across datasets and discussed later in our experiment details. 

\medskip
\noindent \textbf{Inference Details}. For fixed length inputs (ActivityNet-1.3), we fed the full sequence into our model. For variable length inputs (THUMOS14 and EPIC-Kitchens 100), we sent the full sequence into the model. When using position embeddings in our ablation study, we adopted the technique from~\cite{vitarxiv2020}. Specifically, for input sequences shorter than the training sequence length (2304), we fed the full sequence into our model and clipped the position embedding using the actual length of the video. For input sequences longer than the training sequence length, we again fed the full sequence into our model, yet used linear interpolation to upsample the position embeddings.

\medskip
\noindent \textbf{Score Fusion}. For our experiments on THUMOS14 and ActivityNet-1.3, we sometimes consider score fusion using external classification scores. Specifically, given an input video, the top-2 video-level classes given by external classification scores were assigned to all detected action instances in this video, where the action scores from our model were multiplied with the external classification scores. Each detected action instance from our model thus creates two action instances. We refer the readers to~\cite{pgcniccv2019} (Appendix E) for a more detailed description of the score fusion strategy.

\medskip
\noindent \textbf{Experiment Details}.
Our experiment details vary across datasets, as each dataset includes videos of different resolution and frame rate, and considers different types of features. We now describe our experiment details for THUMOS14, ActivityNet-1.3, and EPIC-Kitchens 100.\smallskip

\squishlist
\item \textbf{THUMOS14}: We used two-stream I3D~\cite{kinetics2017cvpr} pretrained on Kinetics to extract the video features on THUMOS14, following~\cite{cmcscvpr2019,bottumuptaleccv2020}. We fed 16 consecutive frames as the input to I3D, used a sliding window with stride 4 and extracted 1024-D features before the last fully connected layer. The two-stream features were further concatenated (2048-D) as the input to our model. \textit{m}AP@$[0.3$:$0.1$:$0.7]$ was used to evaluate our model. Our model was trained for 50 epochs with a linear warmup of 5 epochs. The initial learning rate was 1e-4 and a cosine learning rate decay is used. The mini-batch size was 2, and a weight decay of 1e-4 was used.

\item \textbf{ActivityNet-1.3}: We used two-stream I3D~\cite{kinetics2017cvpr} and TSP~\cite{tspiccvworkshop2021} for feature extraction, and increased the stride of the sliding window to 16. Following \cite{bsneccv2018,bmniccv2019,gtadcvpr2020}, the extracted features were downsampled into a fixed length of 160 and 192 using linear interpolation for I3D and TSP features, respectively. For evaluation, we used \textit{m}AP@$[0.5$:$0.05$:$0.95]$ and also reported the average \textit{m}AP. Our model was trained for 15 epochs with a linear warmup of 5 epochs. The learning rate was 1e-3, the mini-batch size was 16, and the weight decay was 1e-4. For ActivityNet, we find it is helpful to train our model to generate proposals by considering all actions from a single category, and then use external classification scores for the recognition. This strategy was also used in previous single-stage TAL methods~\cite{afsdcvpr2021}. 

\item \textbf{EPIC-Kitchens 100}: We used a SlowFast network~\cite{slowfasticcv2019} pre-trained on EPIC-Kitchens for feature extraction. This model is provided by~\cite{epicarxiv2020}. We fed 32 frame window with a stride of 16 to extract 2304-D features. Our model was trained on the training set and evaluated on the validation set. A window size of 9 was used for local self-attention. For evaluation, we used \textit{m}AP@$[0.1$:$0.1$:$0.5]$ and report the average \textit{m}AP following~\cite{epicarxiv2020}. Our model was trained for 30 epochs with learning rate 1e-4, mini-batch size 2, and weight decay of 1e-4. 
\squishend

\begin{figure*}[t]
\begin{center}
    \minipage{0.48\textwidth}
      \includegraphics[width=\linewidth]{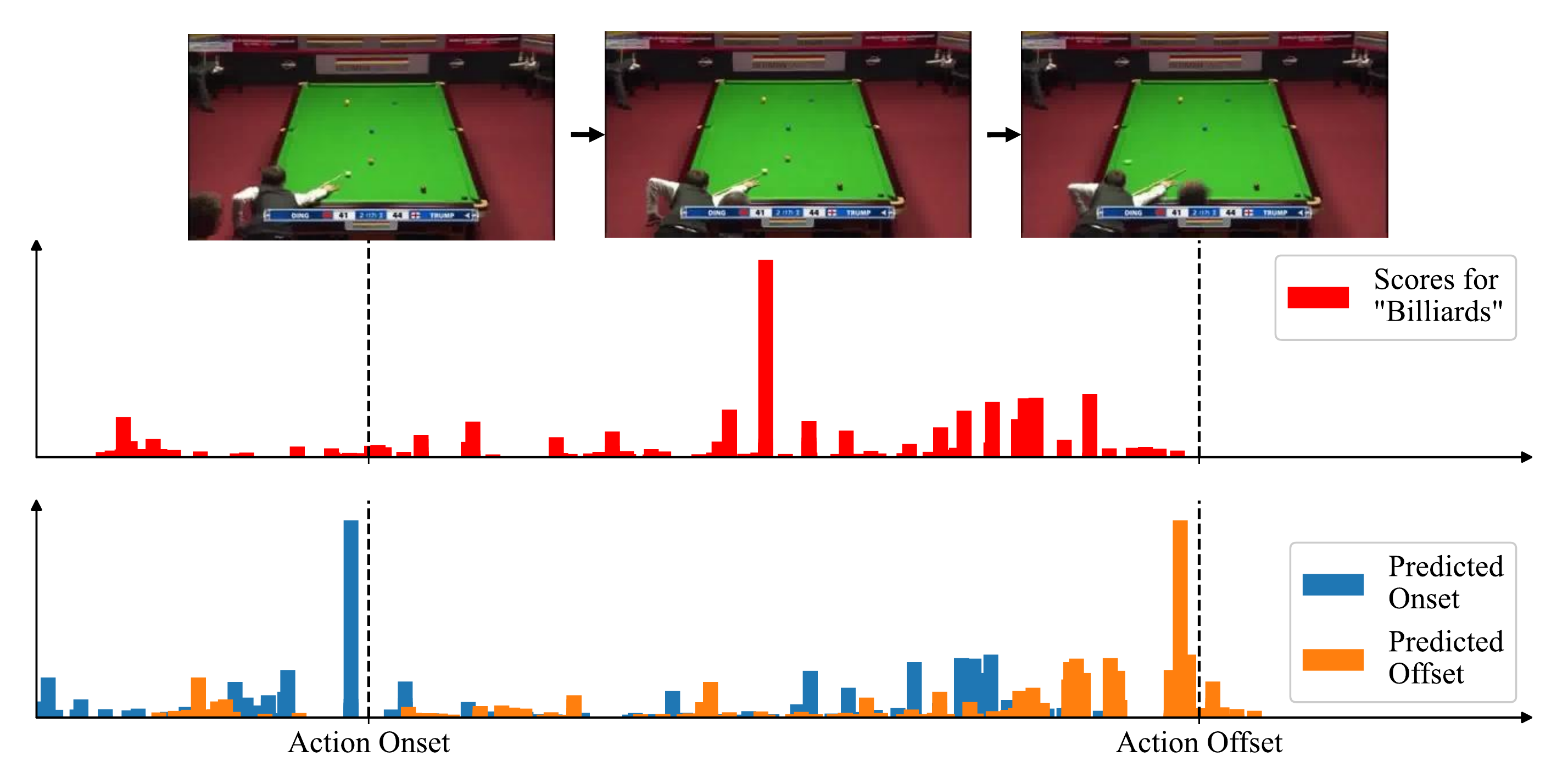}
    \endminipage\hfill
    \minipage{0.48\textwidth}%
      \includegraphics[width=\linewidth]{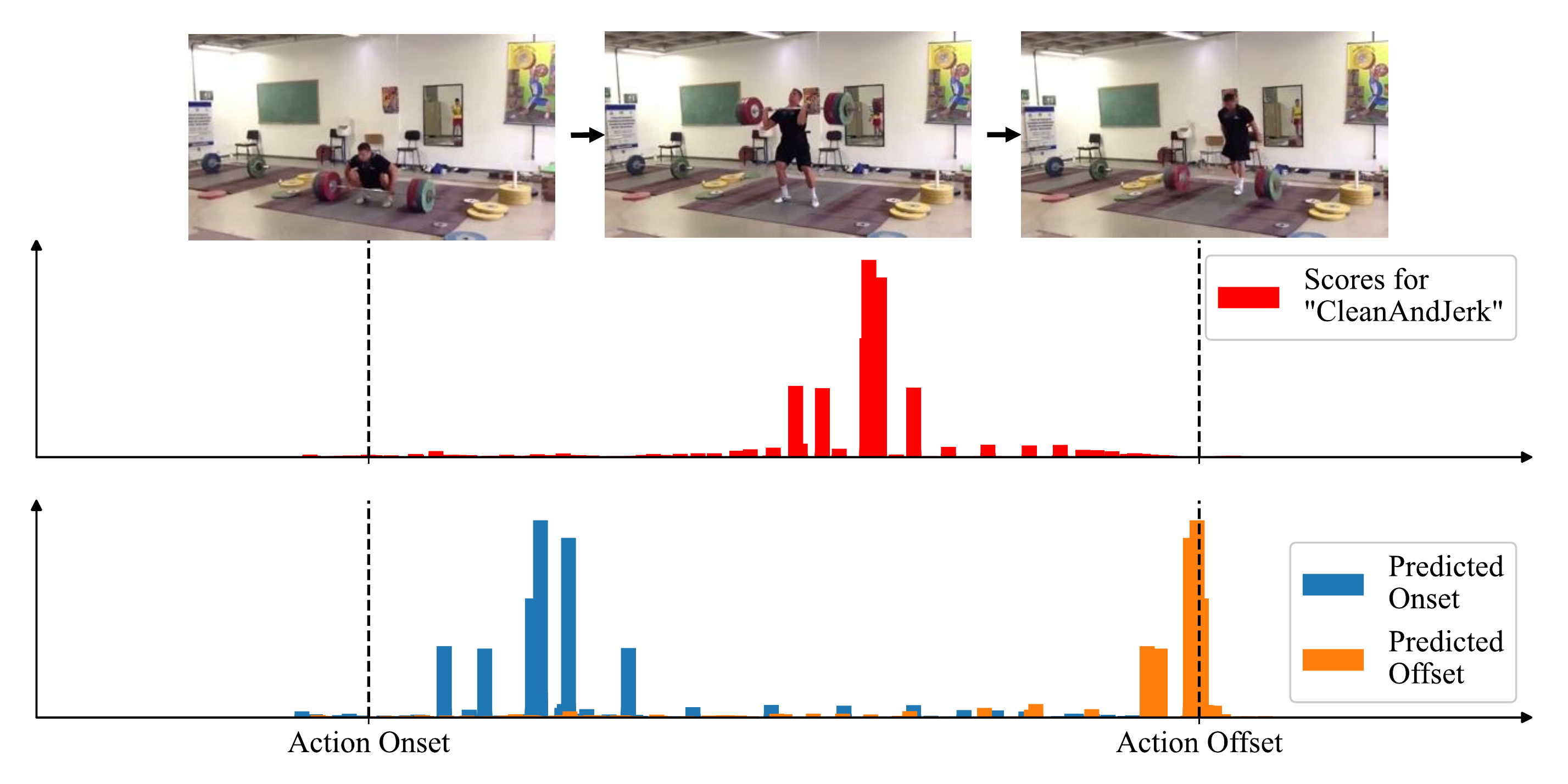}
    \endminipage \\
    \minipage{0.48\textwidth}
      \includegraphics[width=\linewidth]{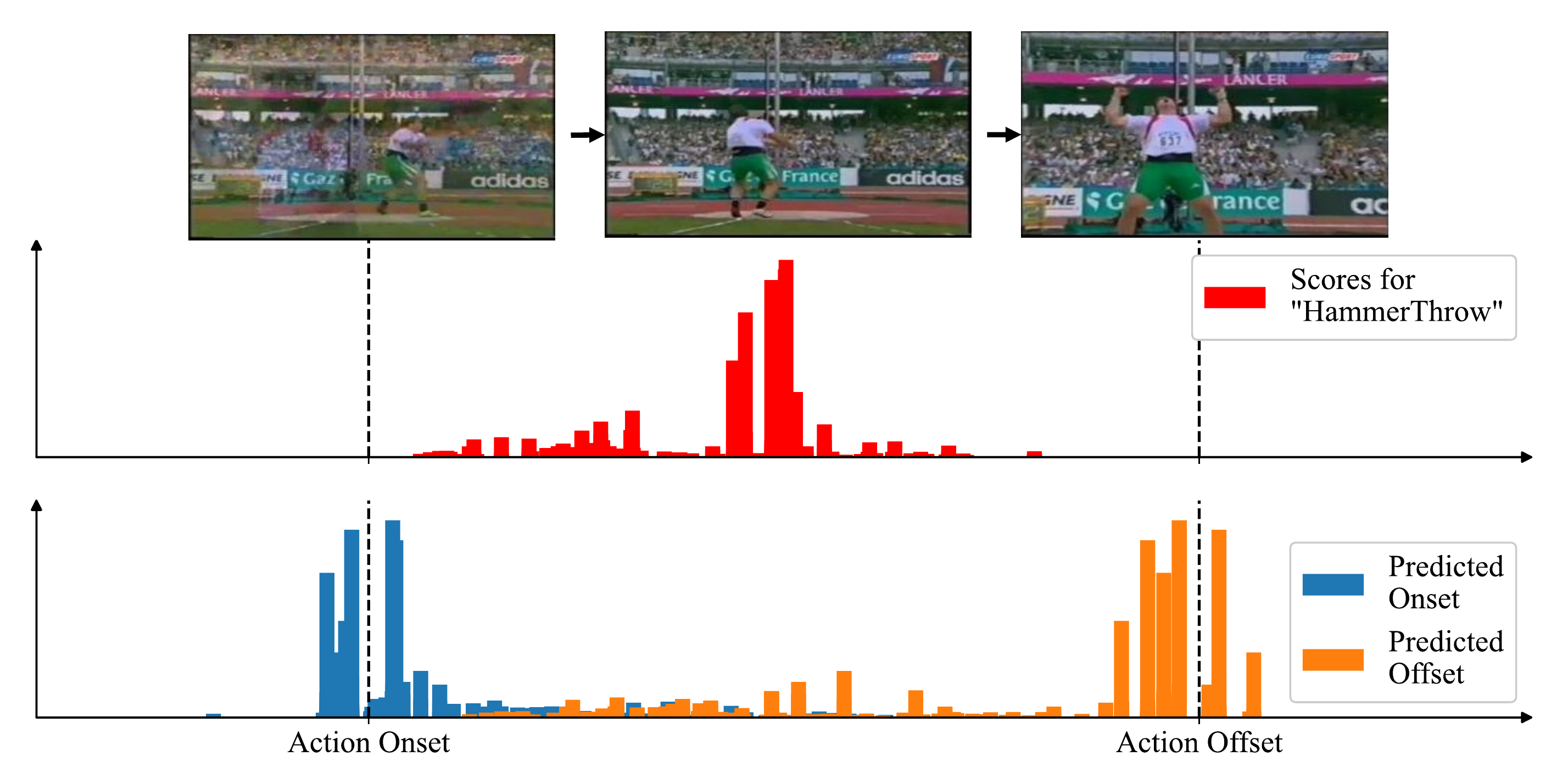}
    \endminipage\hfill
    \minipage{0.48\textwidth}%
      \includegraphics[width=\linewidth]{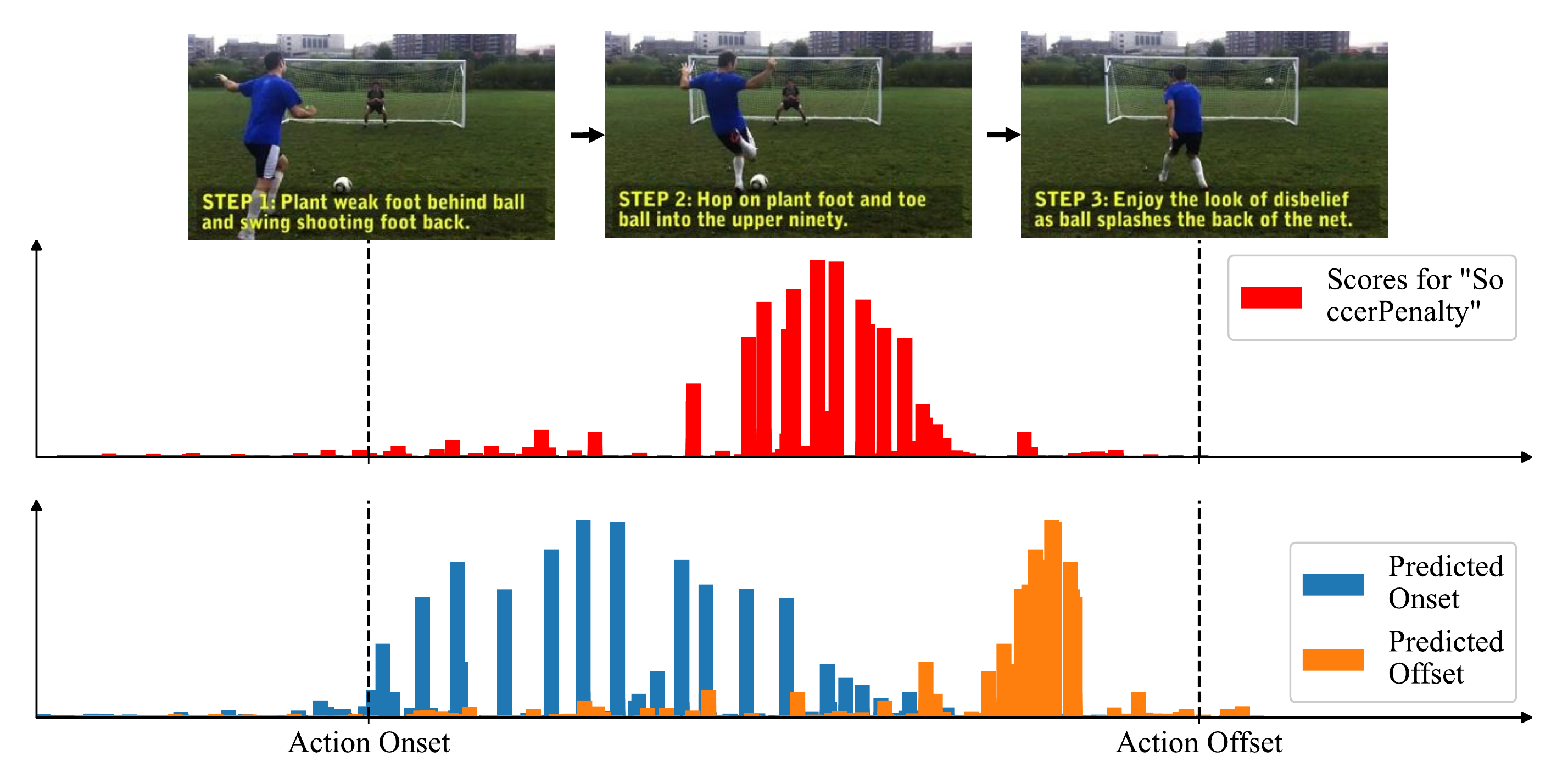}
    \endminipage\\
    \minipage{0.48\textwidth}
      \includegraphics[width=\linewidth]{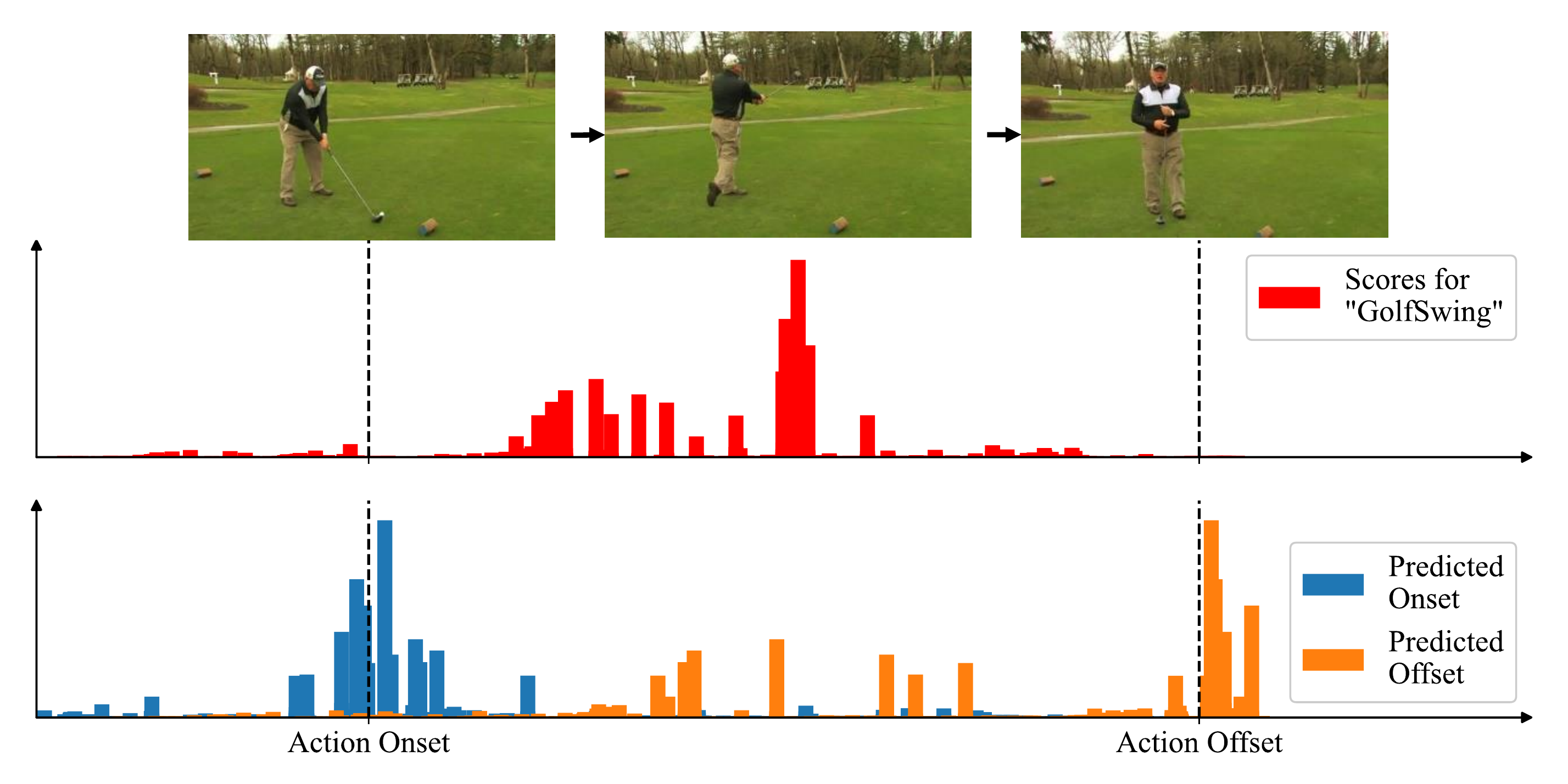}
    \endminipage\hfill
    \minipage{0.48\textwidth}%
      \includegraphics[width=\linewidth]{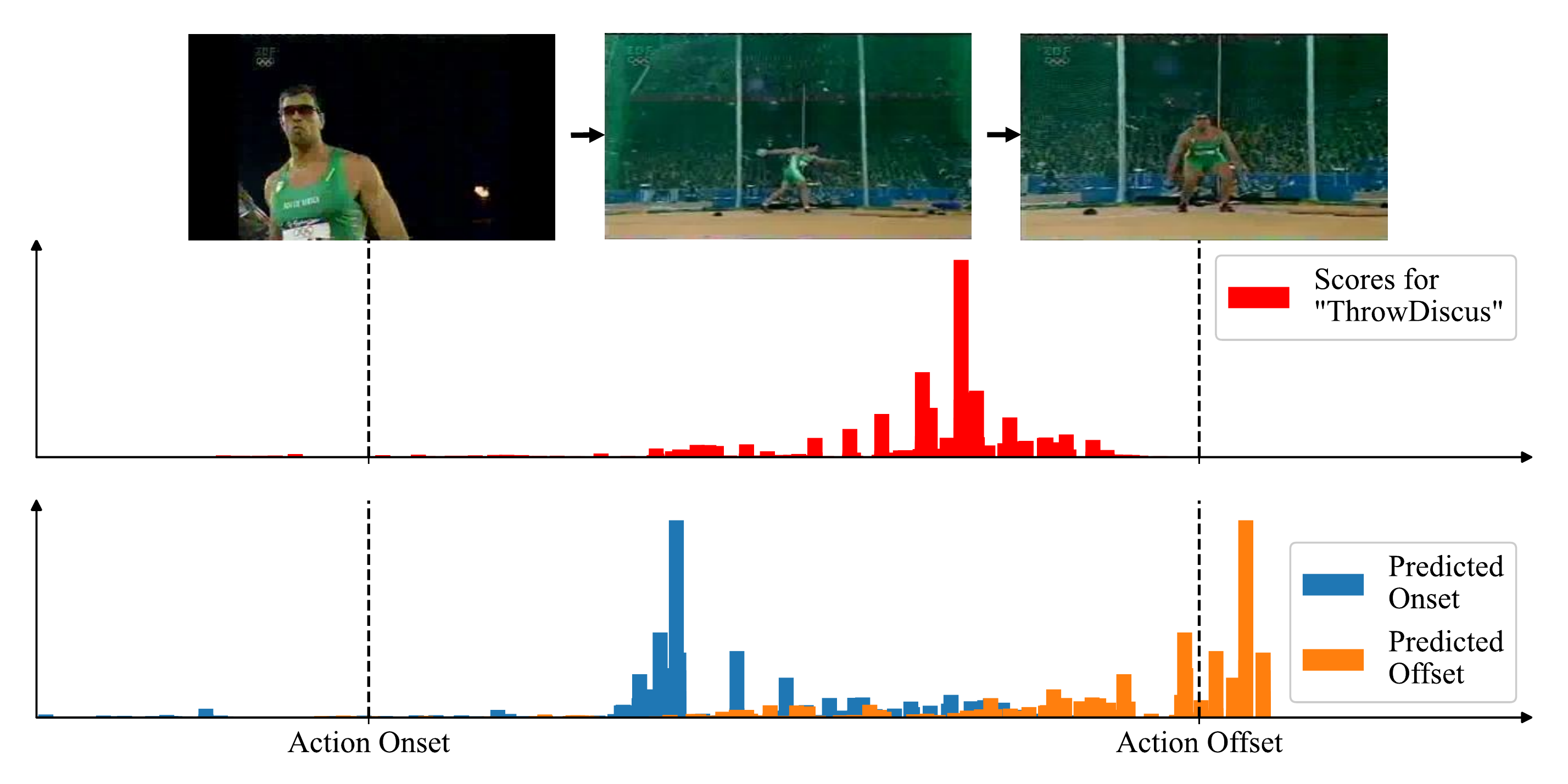}
    \endminipage
\end{center}
\caption{More visualization of our outputs. For each item from \emph{top} to \emph{bottom}: (1) the input video frames;  (2) action scores at each time step; (3) histogram of action onsets and offsets computed by weighting the regression outputs using the action scores. \emph{Left}: successful cases; \emph{Right}: failure cases. This figure is best viewed in color and when zoomed in.}
\label{fig:more vis}
\end{figure*}

\medskip
\noindent \textbf{Reproducibility of Our Results}. All results reported in the paper were obtained with the same random seed using PyTorch 1.10, CUDA 10.2 and CUDNN 7.6.5 on an NVIDIA Titan Xp GPU, using deterministic GPU computing routines. On the same machine, our code will always produce the same results when using the same random seed. Across machines/GPUs and computing environments, we have observed minor variation of average mAP scores (up to 0.5\% average mAP on THUMOS, less than 0.2\% average mAP on ActivityNet, and under 0.8\% average mAP on EPIC Kitchens), yet those minor variations do not erode the clear performance gains of our method. Our code is made publicly available.\medskip

\section{Additional Visualizations}\label{sec:appendix:vis}
Further, we present more visualizations of our results in Fig.\ \ref{fig:more vis}, extending Fig.\ 3 of the main paper. Our model is able to detect the occurrence of actions and estimate their temporal boundaries for the most of the cases (see the first column of Fig.\ \ref{fig:more vis}). The major failure modes of our model, as demonstrated in the second column of Fig.\ \ref{fig:more vis}, include (1) incorrect classification of action centers, \ie background confusion (classification errors); (2) inaccurate regression of the action's onset and offset (localization errors). We plan to address these issues in our future work.

\section{Limitations and Future Work}\label{sec:appendix:limit}
A main limitation of our method is the use of pre-extracted video features, also faced by many previous approaches. Another limitation is the need for many human labeled videos for training and the constraint of a pre-defined vocabulary of actions. Interesting future directions include pre-training for action localization~\cite{tspiccvworkshop2021,xu2021boundary}, and learning from videos and text corpus~\cite{radford2021learning,jia2021scaling} without human labels.

\bibliographystyle{splncs04}
\bibliography{egbib}
\end{document}